\PassOptionsToPackage{numbers, compress}{natbib}
\documentclass[10pt, logo]{nvidiatechreport}

\usepackage{natbib}

\usepackage[utf8]{inputenc} 
\usepackage[T1]{fontenc}    
\usepackage{hyperref}       
\usepackage{url}            
\usepackage{booktabs}       
\usepackage{amsfonts}       
\usepackage{nicefrac}       
\usepackage{microtype}      
\usepackage{xcolor}         

\usepackage{graphicx}
\usepackage{subcaption}
\usepackage{hyperref}
\usepackage{amsmath}
\usepackage{amssymb}
\usepackage{mathtools}
\usepackage{amsthm}
\usepackage{amsmath} 
\usepackage{xcolor}
\usepackage{xparse}

\newcommand{\nips}[1]{{{#1}}}%
\newcommand{\camera}[1]{{{#1}}}

\newcommand{\third}[1]{\underline{#1}}

\newcommand{\cifart}{\text{CIFAR-10}}
\newcommand{\cifarh}{\text{CIFAR-100}}
\newcommand{\femnist}{\text{FEMNIST}}
\newcommand{\news}{\text{20NewsGroups}}

\newcommand{\ours}{\text{FedRevive}}
\newcommand{\sync}{\text{FedAvg}}

\newcommand{\fedbuff}{\text{FedBuff}}
\newcommand{\kdrevive}{\text{KD-Revive}}
\newcommand{\asyncdw}{\text{AFL-DW}}
\newcommand{\oursdd}{\text{DD-FedRevive}}
\newcommand{\fedfa}{\text{FedFa}}
\newcommand{\fadas}{\text{FADAS}}
\newcommand{\fedac}{\text{FedAC}}

\newcommand{\apdx}{\text{Appendix}}
\newcommand{\sect}{\text{Section}}

\NewDocumentCommand{\x}{ O{} O{} }{%
  \mathbf{x}_{\smash{ #1}}^{\smash{ #2}}%
}

\NewDocumentCommand{\tx}{ O{} O{} }{%
  \mathbf{\widetilde{x}}_{\smash{ #1}}^{\smash{ #2}}%
}

\NewDocumentCommand{\upd}{ O{} O{} }{%
  \mathbf{\Delta}_{\smash{ #1}}^{\smash{ #2}}%
}

\NewDocumentCommand{\fit}{ O{} O{} }{%
  f_{\smash{ #1}}^{\smash{ #2}}%
}

\newcommand{\supp}[1]{{ (#1)}}

\NewDocumentCommand{\tauit}{ O{i} O{t} }{
  \tau_{\smash{ #1}}^{\smash{ #2}}
}

\newcommand{\genp}{\boldsymbol{\theta}}
\NewDocumentCommand{\gnrtr}{ O{\genp} }{
  G_{#1}
}

\newcommand{\noise}{\mathbf{z}}
\NewDocumentCommand{\nlabel}{ O{\noise} }{
  \mathbf{y}_{#1}
}

\newcommand{\lr}{\eta}
\newcommand{\lrc}{\lr_c}
\newcommand{\lrs}{\lr_s}
\newcommand{\locit}{K}
\newcommand{\mybeta}{\beta}
\newcommand{\Ncl}{N}

\newcommand{\bs}{B}

\newcommand{\kdbuff}{\mathcal{C}}
\newcommand{\kdbs}{c}

\newcommand{\kddata}{\mathcal{D}_\text{KD}}

\newcommand{\ksynth}{K_\text{synth}}
\newcommand{\Lsynth}{\mathcal{L}_\text{synth}}
\newcommand{\Ladv}{\mathcal{L}_\text{adv}}
\newcommand{\Lfeature}{\mathcal{L}_\text{feature}}
\newcommand{\Ltarget}{\mathcal{L}_\text{target}}
\newcommand{\Lkd}{\mathcal{L}_\text{KD}}

\newcommand{\kkd}{K_\text{KD}}

\newcommand{\tgen}{T_\text{gen}}

\newcommand{\clp}{\hat{\boldsymbol{\pi}}}
\newcommand{\clpi}{\hat{\boldsymbol{\pi}}_i}
\newcommand{\clpij}{\hat{\pi}_i^j}
\newcommand{\clpkj}{\hat{\pi}_k^j}

\newcommand{\nclp}{\hat{\boldsymbol{\omega}}}

\newcommand{\nclpij}{\hat{\omega}_i^j}

\newcommand{\softm}{\sigma}
\usepackage{enumitem}
\usepackage{empheq}
\usepackage{tcolorbox}
\usepackage{nccmath}
\usepackage{caption}
\usepackage{algorithm}
\usepackage{algorithmic}

\makeatletter

\newcommand{\alglinelabel}[1]{%
  \begingroup
    \edef\@currentlabel{\number\value{ALC@line}}%
    \label{#1}%
  \endgroup
}
\makeatother

\newcommand{\Statex}{\item[]}

\usepackage[capitalize,noabbrev]{cleveref}
\Crefname{figure}{Fig.}{Figs.}
\usepackage{wrapfig}

\renewcommand{\monthyeardate}{\renewcommand{\today}{}}
\makeatletter\g@addto@macro\ps@firststyle{}\makeatother

\title{Reviving Stale Updates: Data-Free Knowledge Distillation for Asynchronous Federated Learning
}

\makeatletter
\newcommand{\blfootnote}[1]{%
  \begingroup
  \renewcommand\@makefnmark{}%
  \footnotetext{\vspace{-0.8em}\small #1}%
  \endgroup
}
\makeatother
\usepackage{fontawesome5}

\author{Baris Askin\textsuperscript{1,$\dagger$} \qquad Holger R.~Roth\textsuperscript{2} \qquad Zhenyu Sun\textsuperscript{3,$\dagger$} \qquad Carlee Joe-Wong\textsuperscript{1} \qquad Gauri Joshi\textsuperscript{1} \qquad Ziyue Xu\textsuperscript{2} \\~\\
\textsuperscript{1}Carnegie Mellon University \quad \textsuperscript{2}NVIDIA \quad \textsuperscript{3}Northwestern University
}

\begin{document}

\begin{abstract}
\textbf{Abstract:} \hspace{2pt}Federated learning (FL) enables collaborative model training across distributed clients without sharing raw data, yet its scalability is limited by synchronization overhead. Asynchronous federated learning (AFL) alleviates this issue by allowing clients to communicate independently, thereby improving wall-clock efficiency in large-scale, hardware-heterogeneous environments. However, asynchrony introduces updates computed on outdated global models (\emph{staleness}) that can destabilize optimization and hinder convergence. We propose \ours{}, an AFL framework that revives stale updates through data-free knowledge distillation (DFKD). \ours{} integrates parameter-space aggregation with a server-side DFKD process that transfers knowledge from stale client updates to the current global model without access to data. A meta-learned generator synthesizes pseudo-samples for multi-teacher distillation. A hybrid aggregation scheme combining raw with DFKD updates effectively mitigates staleness while retaining AFL scalability. Experiments on vision and text benchmarks show that \ours{} achieves faster training by up to 29.7\% and higher final accuracy by up to 11.4 percentage points than baselines.
\end{abstract}

\maketitle
\enlargethispage{2\baselineskip}

\blfootnote{%

Accepted at the 40th Annual Conference on Neural Information Processing Systems (NeurIPS 2026).

$^{\dagger}$ Work was done during an internship at NVIDIA.

Correspondence to: Baris Askin \texttt{<baskin@andrew.cmu.edu>}, Holger Roth \texttt{<hroth@nvidia.com>}, Ziyue Xu \texttt{<ziyuex@nvidia.com>}.

}

\begin{center}
\small
\href{https://nvidia.github.io/NVFlare/research/fedrevive}{%
  \raisebox{-0.15\height}{%
    \includegraphics[height=1.1em]{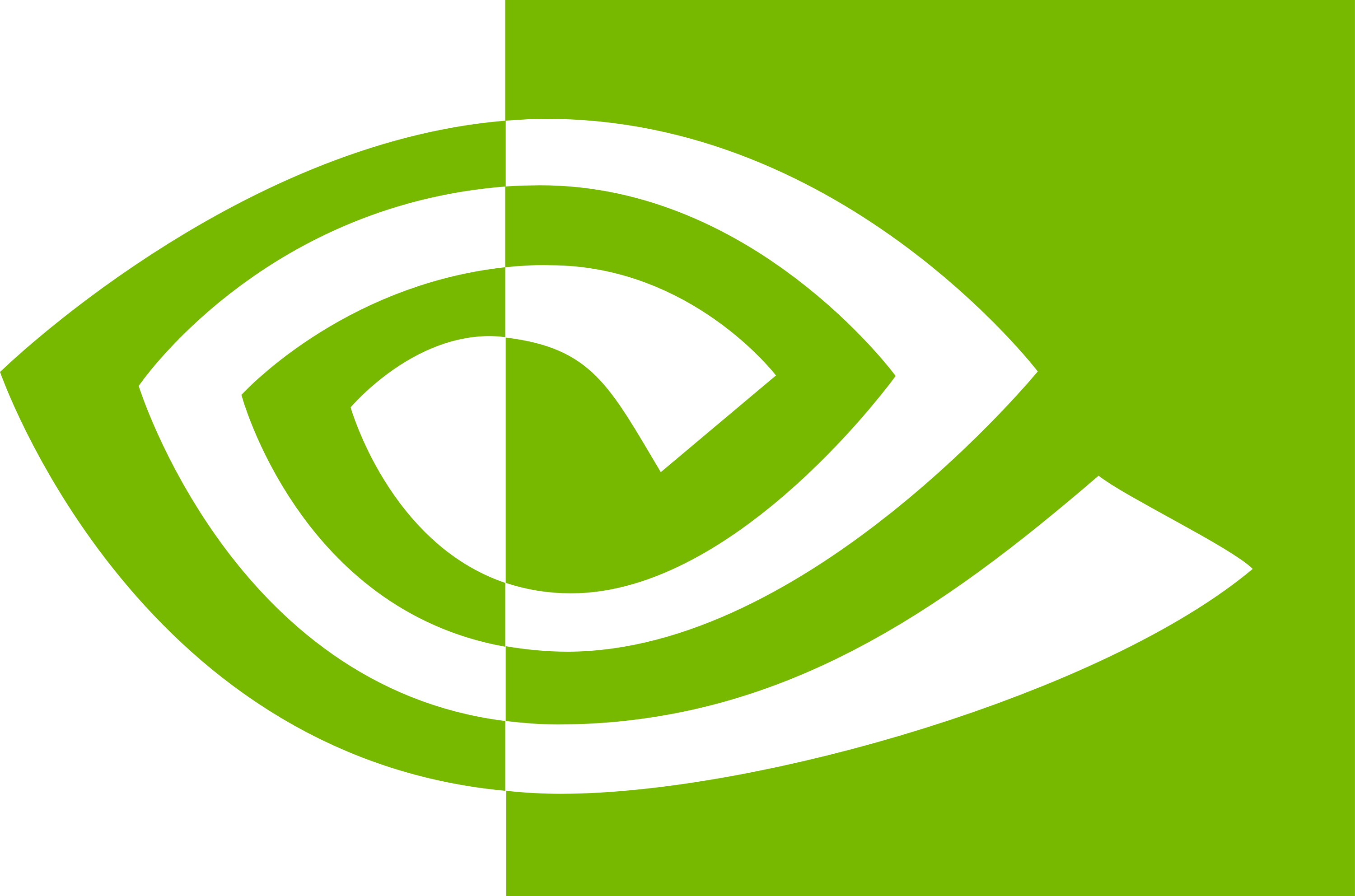}}%
  \enspace NVIDIA FLARE Implementation%
}
\qquad
\href{https://github.com/askinb/FedRevive}{%
  \faGithub\enspace Project Codes%
}
\end{center}

\section{Introduction}
\label{intro}

Modern artificial intelligence models thrive on large amounts of data, which raises serious concerns about privacy and regulatory compliance~\citep{yang2019federated}. Federated learning (FL) can help address this dilemma by enabling many clients (e.g., smartphones, wearables, hospital servers) to collaboratively train a shared model without uploading raw data \cite{FedAvg}. In vanilla FL, the server broadcasts the current global model, clients perform local training on private data, and the server aggregates returned updates to form the next global iterate. This paradigm preserves data locality and widens access to otherwise restricted data, while reducing central storage risks. Moreover, relying only on public datasets is increasingly insufficient, as large models saturate public corpora and demand broader coverage across domains. FL supports both cross-silo (e.g., hospitals and enterprises) and cross-device (e.g., millions of phones) settings, offering a practical path to leverage diverse private data at scale~\citep{rieke2020future,FedScale}.

In \emph{synchronous} FL, the server waits for all (or a selected subset of) participating clients to complete their local updates before aggregation~\citep{FedAvg}. Although this strategy aligns updates to the same global state, it suffers from the \emph{straggler problem}: slow or unavailable clients delay the entire round, leading to underutilized resources and poor scalability under heterogeneous availability and compute capacity~\citep{xie2019asynchronous}.  \emph{Asynchronous} FL (AFL) alleviates this bottleneck by allowing the server to incorporate each client update upon arrival, without waiting for others~\citep{xie2019asynchronous,FedBuff}. Clients train and communicate independently, and the server continuously updates the global model. This design improves wall-clock efficiency and makes AFL appealing for cross-device deployments such as mobile networks and for cross-silo collaborations where institutional availability fluctuates.

The flexibility of AFL, however, comes with a major drawback: \emph{update staleness}. Since clients may train on older global checkpoints, the server receives delayed updates that no longer match the current optimization state, which can distort training dynamics and degrade performance, especially under not independently and identically distributed (non-IID) data~\citep{xie2019asynchronous,FedBuff,FedAST,richtarik2025handling}. This issue becomes more pronounced as the system scales, since larger deployments naturally exhibit a wider spread of client runtimes and delays~\citep{koloskova2022sharper}. Existing AFL methods mitigate staleness by buffering or smoothing arriving updates~\citep{FedBuff} or by explicitly correcting or down-weighting delayed updates~\citep{chen2021fedsa}. While effective in certain regimes, these strategies can slow server progress or suppress information carried by slower clients. Even a delayed update can encode rich signals about a client’s local data. The key challenge is therefore not only to control the harmful drift induced by staleness, but also to retain the useful knowledge in delayed client models. This motivates the following question:

\vspace{-0.5em}
\begin{center}
\textit{Can we \textbf{revive} stale updates in asynchronous FL to extract useful knowledge for improved scalability?}
\end{center}
\vspace{-1em}

\paragraph{Our idea: distill knowledge from stale updates.}
We answer this question by combining parameter-space aggregation with \emph{data-free knowledge distillation} (DFKD). Knowledge distillation (KD) transfers information from one or more teachers to a student via softened targets or intermediate representations~\citep{KD}. Data-free variants remove the assumption that a public dataset is available for distillation~\citep{first_dfkd}, which aligns with practical FL deployments. At a high level, our method treats received client models as teachers and the current server model as the student, and uses server-side DFKD to extract transferable knowledge from delayed client states. Then, by coupling the regular parameter-space update with a DFKD update, our method mitigates the adverse effects of staleness. Integrating DFKD into AFL is nontrivial because (i) no data are typically available on the server, (ii) stale client models can be heterogeneous and \nips{only partially trained, and (iii) the additional server-side processing should remain moderate. 
\ours{} targets cross-device FL regimes where stale updates due to a large number of clients harm optimization.
}

\paragraph{Related work.}
\camera{AFL builds on earlier work on asynchronous distributed stochastic gradient descent~\citep{dean2012large,lian2015asynchronous}. In federated settings, asynchronous aggregation has been explored as a practical remedy for stragglers by integrating updates upon arrival rather than at synchronized rounds~\citep{xie2019asynchronous,FedBuff,XU2023100595,online_afl,richtarik2025handling}.}
To mitigate staleness, prior work explores buffering or delay-aware reweighting strategies that control instability but can also attenuate information from slower clients~\citep{FedBuff,chen2019communication,9407951,9022982,zhang2016staleness,dutta2018slow,cui2014exploiting}. In parallel, knowledge distillation (KD) is a tool for transferring predictive structure from teachers to students~\citep{KD}, and has been adapted to FL to mitigate heterogeneity, support model-architecture mismatch, and improve privacy via logits or ensemble-based supervision~\citep{li2019fedmdheterogenousfederatedlearning,lin2020ensemble,qin2024knowledgedistillationfederatedlearning,10825769}. 
\camera{
These FL-KD methods typically assume access to a shared dataset for distillation, either on the server or on clients~\citep{fl_kd_server1,itahara2021distillation,li2026fedcd}. DFKD removes this dependency by synthesizing pseudo-samples for distillation without real data~\citep{first_dfkd,mansourian2025a,ReptileKD}, and several recent works adopt DFKD within synchronous FL to avoid public data while improving robustness under non-IID data~\citep{gao2025feddtgfederateddatafreeknowledgedistillation,zhang2022fine,zhu2021data,wang2023dafkd}.
}
However, the use of KD in {asynchronous} FL remains largely unexplored. Some efforts rely on public data assumptions often unrealistic in FL deployments~\citep{lu2025correctedlatestversionmake}. Unlike synchronous FL, where distillation leverages relatively stable teacher models, AFL distills from client models misaligned with the current server state. 
\camera{
Although \ours{} builds on established DFKD techniques, it introduces an AFL-specific update that distills predictive knowledge from stale client models into the current server model. Adapting distillation to continually changing, partially trained, and heterogeneous teachers poses nontrivial design challenges, which we carefully address through a persistent meta-learned generator, class-aware multi-teacher distillation, and staleness-dependent hybrid aggregation.}
\apdx~\ref{appndx:extended_related_work} presents an extended related work discussion.

\paragraph{Contributions.}
The main contributions of the paper are summarized as follows:
\begin{itemize}[
    wide=0pt,
   topsep=1pt,
    itemsep=1pt,
    parsep=0pt,
    partopsep=0pt
]
\item We propose \ours{}, an AFL framework that \emph{revives} stale client updates by combining parameter aggregation with server-side, multi-teacher DFKD.
\item We carefully design a DFKD pipeline tailored to AFL without requiring any extra client metadata beyond regular FL updates. We leverage multi-teacher distillation with computation-light process to operationalize our idea.
\item We demonstrate improved speed and accuracy over the baseline methods under realistic simulations with image and text-domain experiments, achieving faster time-to-target accuracy and better final performance, and support the proposed method with ablation studies \nips{on hyperparameters, algorithmic components and overhead analysis.}
\end{itemize}

Our results indicate that stale client updates still carry rich, transferable knowledge. By extracting that knowledge in a data-free manner and fusing it with parameter-space aggregation, AFL can retain its scalability advantages without sacrificing model quality.

\section{FL Problem \& Proposed Method}
\label{sect:method}

\begin{wrapfigure}[22]{r}{0.57\textwidth}
  \vspace{-2.5\baselineskip}
  \begin{minipage}{\linewidth}\scriptsize
\begin{algorithm}[H]
\caption{Generic AFL Framework}
\label{alg:afl_framework}
\begin{algorithmic}[1]
\Statex \textbf{Server Process:}
\STATE \textbf{Inputs:} Initial model $\x[][\supp{0}]$, server and  client learning rates $\lrs$ and $\lrc$, \#local iterations $\locit$, \#rounds $T$, 
\#selected active clients $N_a$.
\STATE \alglinelabel{line:init}Send $\x[][\supp{0}]$ to $N_a$ randomly selected active clients.
\FOR{$t = 0$ \textbf{to} $T-1$}
    \STATE \alglinelabel{line:wait}\textbf{Wait} until the server receives a $\tauit$-stale updated model $\tx[i][\supp{t-\tauit}]$ from client $i$.
    \STATE Client update $\upd[i][\supp{t-\tauit}] \leftarrow \tx[i][\supp{t-\tauit}] - \x[][{\supp{t-\tauit}}]$. \label{line:calc_delta}
\STATE \alglinelabel{line:agg}
\colorbox{blue!10}{\textbf{Process update using one of Eqs.~(\ref{eq:afl_aggr}), (\ref{eq:fedbuff_aggr}), or (\ref{eq:ours_aggr}).}}\hspace{-5pt}
    \STATE \alglinelabel{line:send} Assign local training to a randomly selected active client with the updated model $\x[][\supp{t+1}]$. 
\ENDFOR 
\vspace{0.4em}
\\\textbf{Client $i$'s Local Training (Async. and Parallel):} 
\STATE \textbf{Input:} Received server model $\x[][]$. \label{line:client_start} 
\STATE Initialize $\tx \leftarrow \x$.
\FOR{$k = 1$ \textbf{to} $\locit$}
\STATE $\smash{\tx \leftarrow \tx - \lrc \widetilde{\nabla} \fit[i][](\tx)}$ \COMMENT{A stoch. grad. update}
\ENDFOR
\STATE \textbf{Return} {$\tx$} to the server. \label{line:client_end}
\end{algorithmic}
\end{algorithm}
  \end{minipage}
\end{wrapfigure}

\paragraph{Problem formulation.}
We consider an FL setup with $\Ncl$ clients collaboratively training a global model {$\x\in\mathbb{R}^d$}. 
Each client $i$ has  {$f_i:\mathbb{R}^d\rightarrow\mathbb{R}$}, a local objective function defined on private data. Local objectives are different across clients due to heterogeneity in data distribution.  
The goal of FL is to find $\x$ minimizing the global objective, formulated as the average of local losses, $\frac{1}{\Ncl}\sum_{i\in[\Ncl]} f_i(\x)$.

\paragraph{Asynchronous framework.}
We first describe a generic AFL framework that encompasses both existing asynchronous schemes~\citep{xie2019asynchronous, FedBuff} and our proposed method, $\ours$. The overall process is summarized in Algorithm~\ref{alg:afl_framework}. After model initialization (line~\ref{line:init}), the server sends $\x[][\supp{0}]$ to $N_a$ randomly selected active clients for local training. Then, for each server step $t$, the server waits for the next returned model $\tx[i][\supp{t-\tauit}]$ from some client $i$ (line~\ref{line:wait}), where $\tauit$ denotes the staleness, namely, how many server updates occurred since the client was asked to perform local training. The server 
processes the client update via an aggregation rule (line~\ref{line:agg}), and dispatches the updated model $\x[][\supp{t+1}]$ to a randomly selected active client (line~\ref{line:send}) to sustain parallelism. Different AFL variants share this control flow and differ only in the aggregation step, that is, how the server maps $(\x[][\supp{t}], \upd[i][\supp{t-\tauit}])$ to $\x[][\supp{t+1}]$.

\paragraph{Vanilla and buffered AFL.}
Vanilla AFL algorithm~\cite{xie2019asynchronous} updates the global model immediately when a client update arrives. Accordingly, the aggregation step in Algorithm~\ref{alg:afl_framework} (line~\ref{line:agg}) becomes:

{
\noindent\centering
\colorbox{blue!5}{%
  \parbox{0.99\linewidth}{%
    \centering
    \setlength{\abovedisplayskip}{6pt}
    \setlength{\belowdisplayskip}{3pt}
    \setlength{\abovedisplayshortskip}{3pt}
    \setlength{\belowdisplayshortskip}{3pt}
    \begin{align}
      \x[][\supp{t+1}] &\leftarrow \x[][\supp{t}] + \lrs\, \upd[i][\supp{t-\tauit}] \label{eq:afl_aggr}
    \end{align}
  }%
}
}

While this rule is simple and maximally asynchronous, it makes staleness accumulate fast. Every time the server updates $\x$, all clients that are still training are effectively one step further behind. When staleness, $\tauit$, is large, the returned update can yield a biased step and lead to oscillations or divergence.
\fedbuff{}~\citep{FedBuff} mitigates this by buffering updates before applying them. The server maintains a buffer $\mathcal{B}$ of size $\bs$, stores incoming client updates, and aggregates only when $\bs$ updates have been collected. The aggregation step in Algorithm~\ref{alg:afl_framework} (line~\ref{line:agg}) becomes:

{
\noindent\centering
\colorbox{blue!5}{%
  \parbox{0.99\linewidth}{%
    \centering
    \setlength{\abovedisplayskip}{6pt}
    \setlength{\belowdisplayskip}{3pt}
    \setlength{\abovedisplayshortskip}{3pt}
    \setlength{\belowdisplayshortskip}{3pt}
    \begin{align}
      &\mathcal{B} \leftarrow \mathcal{B} \cup \{\upd[i][\supp{t-\tauit}]\} \nonumber\\
      \textbf{if } |\mathcal{B}| = \bs: \; \x[][\supp{t+1}] \leftarrow \x[][\supp{t}] +&
      \mfrac{\lrs}{\bs}
      {\textstyle \sum_{\upd \in \mathcal{B}}} \upd \hfill
      \;\text{ and }\;\mathcal{B} \leftarrow \emptyset, \;\;\;\;
      \textbf{   else: }\; \x[][\supp{t+1}] \leftarrow \x[][\supp{t}] \label{eq:fedbuff_aggr}  
    \end{align}
  }%
}%
}

The buffered strategy smooths the staleness effect by reducing the update frequency.  
Yet, as the buffer size $\bs$ increases, the infrequent server updates slow down the training.
Next, we describe our method, $\ours$, mitigating the staleness without reducing the server model update frequency.

\subsection{Overview of $\ours$}

\mbox{\ours} addresses the staleness problem by leveraging knowledge-distillation. Although directly aggregating the parameters of stale updates can result in training instabilities, the received update still contains informative signals from client data. We augment the standard parameter-space update with a data-free knowledge distillation (DFKD) signal extracted from the updated client model.

\paragraph{$\ours$ aggregation.}

In \mbox{\ours}, the server adaptively fuses the conventional parameter-space update with a knowledge-distilled update to alleviate the adverse effect of staleness.  
When the server receives a $\tauit$-stale client update $\upd[i][\supp{t-\tauit}]$, it first obtains a knowledge-distilled update through a DFKD process.
The server then combines it with original parameter update through an adaptive weighting function $\mybeta(\tauit) \in [0,1]$ controlling the reliance on the knowledge-distilled signal according to the degree of staleness.  
The aggregation step in Algorithm~\ref{alg:afl_framework} (line~\ref{line:agg}) becomes:

{
\noindent\centering
\colorbox{blue!5}{%
  \parbox{0.99\linewidth}{%
    \centering
    \setlength{\abovedisplayskip}{6pt}
    \setlength{\belowdisplayskip}{3pt}
    \setlength{\abovedisplayshortskip}{3pt}
    \setlength{\belowdisplayshortskip}{3pt}
\begin{align}
    &\upd[i][\text{KD}] \leftarrow \kdrevive\!\left(\tx[i][\supp{t-\tauit}],\, \x[][\supp{t}]\right),\nonumber\\
    &\x[][\supp{t+1}] \!\leftarrow\! \x[][\supp{t}]
\!+\! \lrs \bigl[(1\!-\!\mybeta(\tauit))\upd[i][\supp{t-\tauit}]
\!+\! \mybeta(\tauit)\upd[i] [\text{KD}]\bigr]\!\! \label{eq:ours_aggr}
\end{align}
  }%
}%
}

where $\kdrevive(\cdot)$ encapsulates our proposed DFKD mechanism.
The function $\mybeta: \mathbb{N}_0\rightarrow[0, 1]$ is a non-decreasing mapping of the staleness $\tauit$, such that the server increasingly relies on the KD-based update as the client update becomes more outdated.  
Intuitively, when $\tauit \approx 0$, the client model is nearly up-to-date, and the raw parameter update remains reliable, so $\mybeta(\tauit)\!\approx\!0$; conversely, when $\tauit$ is large, $\mybeta(\tauit)\!\approx\!1$, distilled knowledge extracted from the stale client model is weighted more.  
The choice of $\mybeta(\tauit)$ is detailed in Section~\ref{sect:experiments}.

\paragraph{Challenges of integrating KD in AFL.}
Integrating KD into AFL is complicated by the absence of shared data at the server, continually changing teacher models, and client heterogeneity. We first provide an overview of the main issues and how $\ours$ addresses them.

\begin{figure}[t]
\centering
\includegraphics[width=\textwidth]{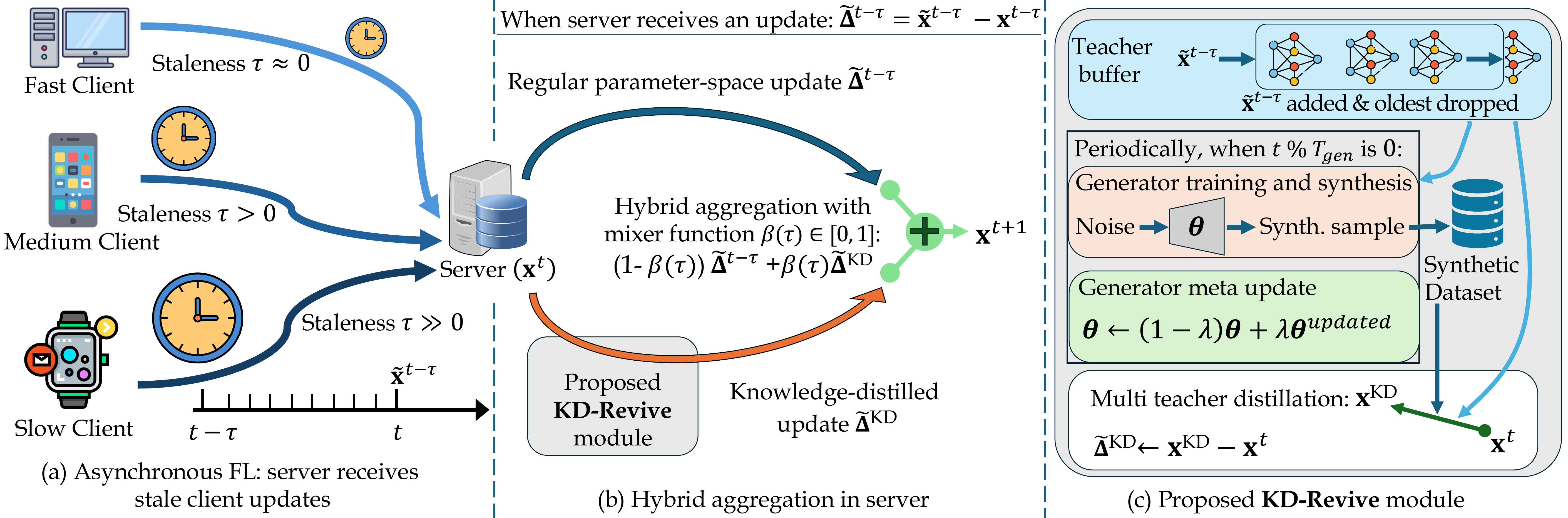}
\caption{\textbf{Proposed \ours{} framework.}}

\label{fig:proposed_framework}
\end{figure}

\textit{(a) Lack of distillation data.}
Standard KD practice requires data to mediate the knowledge transfer between the teacher and the student models. In FL, a public or server-accessible dataset is rarely available due to privacy and communication constraints. 
Our $\kdrevive$ module performs data-free KD and synthesizes pseudo-inputs guided jointly by the incoming client model and the current server model, enabling distillation without access to real data. Note that $\ours$ only requires standard federated updates as in FedAvg \citep{FedAvg} with no additional information from client. Generated pseudo-samples do not provide extra information and are simply intermediate computations for extracting knowledge from already-received updates. 
\nips{Client updates arise after multiple local optimization steps and are mixed with asynchronous delay. Empirically, this does not provide enough signal to synthesize realistic client images, even though it carries intermediate predictive structure useful for knowledge transfer. We provide a discussion and representative synthesized samples in \apdx~\ref{appx:privacy_discussion}.}

\textit{(b) Dynamic teachers and need for lightweight operations.} 
Teacher models in KD or DFKD setups are typically frozen and fully trained oracles. In FL, however, client models continually evolve and are only partially trained. Transferring knowledge from such transient teachers is non-trivial. Moreover, the overhead due to the distillation process must remain computationally light to avoid a new bottleneck. We adopt a lightweight meta-learning strategy inspired by~\cite{ReptileKD}, maintaining a single generator that evolves throughout training. This generator efficiently adapts every $\tgen$ rounds to the incoming client models and synthesizes pseudo-samples to extend the synthetic KD dataset ($\kddata$), initialized empty in the beginning of the training.

\textit{(c) Data heterogeneity across clients.}
Due to limited data size and non-IID distributions, each client’s update tends to overfit to its local data, which leads to biased knowledge transfer. We mitigate this via \textit{multi-teacher distillation}. The server maintains a buffer $\kdbuff$ of the $\kdbs$ most recently received  client models and distills from all buffered teachers jointly. To better exploit client-specific information, the server derives proxy class proportions of clients from the received updates \textit{without additional information}. Our results do not assume access to extra information beyond regular updates.

\paragraph{Per-client class-proportion proxies from regular updates.}

To weight buffered teachers in a class-aware manner, we associate each client $i$ with a \emph{proxy} class-proportion vector $\clpi \in \mathcal{S}^{C}$ where $C$ denotes the number of classes, $\mathcal{S}^{C}$ denotes $C$-simplex, and $\clpij=[\clpi]_j$ denotes the proxy proportion of class $j$ on client $i$.
The server computes $\clpi$ using only the {regular} updates and server-side forward passes, requiring \emph{no additional client computation or information}.

Let $\mathbf{y}_i^{\smash{(u)}}$ denote the $u$-th model upload received from client $i$ (ordered by server reception), and consider the first $T_{\text{est}}$ uploads $\{\mathbf{y}_i^{\smash{(u)}}\}_{u=1}^{T_{\text{est}}}$.
For each upload, the server probes $\mathbf{y}_i^{\smash{(u)}}$ with a batch of random inputs $\mathbf{r}$ (Gaussian noise for vision models, random token sequences for language models), and computes a per-upload proxy:
\[
\widetilde{\boldsymbol{\pi}}_i^{(u)} = 
(1/|\mathbf{r}|){\textstyle\sum_{r\in\mathbf{r}}}
\softm\!\left(\mathbf{y}_i^{\smash{(u)}}(r)\right),
\;\; u=1,\dots,T_{\text{est}},
\]
where $\softm$ denotes the softmax function. The server then averages these vectors and keeps this proxy fixed for the remainder of training,
\(
\clpi \;\gets\; \mfrac{1}{T_{\text{est}}}{\textstyle\sum_{u=1}^{T_{\text{est}}}}\widetilde{\boldsymbol{\pi}}_i^{(u)},
\)
and uses running-average for {\normalsize $u < T_{\text{est}}$}. This adds negligible server overhead since it only requires forward passes and $T_{\text{est}}$ is small (2 in our experiments), while imposing no additional client-side computation.
\camera{Although this proxy is noisy, \ours{} is empirically robust to its errors (\apdx~\ref{appx:estimated_client_distribution}, \Cref{fig:estimated_client_dist_training_curves}). Importantly, it is used only as a coarse dominant-class cue to guide synthesis and teacher weighting, rather than to recover exact client histograms. In \apdx~\ref{appx:estimated_client_distribution}, we report KL divergence of proxy proportions, top-1 and top-2 dominant-class recovery, a controlled study across label-skew levels, and ablation experiments comparing estimated and true class proportions.}

\subsection{Detailed Steps of $\kdrevive$}
\label{subsec:kdrevive_details}
When the server receives an updated model $\tx[i][\supp{t-\tauit}]$ from client $i$ (Algorithm~\ref{alg:afl_framework}, line~\ref{line:wait}), it executes $\kdrevive$ to compute a data-free knowledge-distilled update (Eq.~(\ref{eq:ours_aggr})). This procedure consists of four steps: (i) updating the model buffer, (ii) periodic generator training and synthesis, (iii) meta-update of the generator, and (iv) multi-teacher distillation. \ours{} framework is depicted in \Cref{fig:proposed_framework}.

\paragraph{(1) Update the model buffer.}
The server maintains a KD buffer $\kdbuff$ storing $\kdbs$ most recently received client models to use as teachers. Receiving $\tx[i][\supp{t-\tauit}]$, the buffer is updated as:
\[
\kdbuff \gets \kdbuff  \cup \{\tx[i][\supp{t-\tauit}]\} \backslash \{\text{oldest model in }\kdbuff\}.
\]
The rolling buffer ensures that the distillation process utilizes recent and diverse client knowledge. Let $\mathcal{I}_t$ denote the set of client indices whose models are currently buffered.
For class-aware weighting, we normalize the proxy client proportions across buffered teachers, for each class $j$ and client $i\in\mathcal{I}_t$:
\(
\nclpij \;\triangleq\; {\clpij}/{\sum_{k\in\mathcal{I}_t}\clpkj}
\)
so that $\sum_{i\in\mathcal{I}_t}\nclpij = 1$ for every class $j$.
We use $\nclpij$ to weight teacher contributions in subsequent steps.

\paragraph{(2) Periodic generator training and synthesis.}
A generator $\gnrtr$ with parameters $\genp$ is initialized once at the start of training.
To keep server overhead small, the server runs generator training and synthesis only once every $\tgen$ server updates; in intermediate rounds, it uses the latest $\kddata$ and proceeds directly to Step~(4).
In a synthesis iteration, we sample latent vectors $\noise=\{\noise_b\}_{b=1}^{B}$ and target labels $\nlabel=\{y_b\}_{b=1}^{B}$ (uniformly at random over classes), and train $(\noise,\genp')$ for $\ksynth$ iterations, where $\genp'$ is initialized by $\genp' \gets \genp$:
\[
(\noise, \genp') \gets (\noise, \genp') - \eta_{\text{synth}}\, \nabla_{(\noise, \genp')}\,
\Lsynth\!\big(\gnrtr[\genp'](\noise)\big).
\]
Let $\mathbf{r}_b = \gnrtr[\genp'](\noise_b)$ be a synthetic input, and denote
\mbox{$\tilde{\mathbf{p}}_{k,b}=\softm(\tx[k](\mathbf{r}_b))$} and ${\mathbf{p}}_{b}=\softm(\x[][\supp{t}](\mathbf{r}_b))$ as the  predictive distribution of teacher $k$ in $\kdbuff$ and student (current server model), respectively. The synthesis loss is defined as:
\begin{align}
\Lsynth
= \frac{1}{B}\sum_{b=1}^{B}\sum_{k\in\mathcal{I}_t}\nclp_k^{y_b}\Big[
&\underbrace{\alpha_{\text{target}}\,
\text{Cross-Entropy}\!\big(\tilde{\mathbf{p}}_{k,b},\, y_b\big)}_{\Ltarget\text{: target loss}}
+\underbrace{\alpha_{\text{feature}}\,
\|\phi(\tx[k],\mathbf{r}_b)-\bar{\phi}(\tx[k])\|^2}_{\Lfeature\text{: feature loss}}
\nonumber\\
&-\underbrace{\alpha_{\text{adv}}\,
\mathbf{1}{\{\hat{c}_{k,b}=\hat{c}_{b}\}}\,
\text{KL}\!\big(\tilde{\mathbf{p}}_{k,b}\,\|\,\mathbf{p}_b\big)}_{\Ladv\text{: adversarial loss}}
\Big],
\label{eq:lsynth_combined_twolines}
\end{align}

where $\phi(\tx[k],\mathbf{r}_b)$ extracts intermediate layer statistics from teacher $k$ on input $\mathbf{r}_b$,
and $\bar{\phi}(\tx[k])$ denotes the corresponding running statistics stored in $\tx[k]$. 
The choice of $\phi$ depends on the modality and model architecture. Defining $\hat{c}_{k,b}=\arg\max_j (\tilde{\mathbf{p}}_{k,b})_j$ and $\hat{c}_{b}=\arg\max_j ({\mathbf{p}}_b)_j$, $\Ladv$ discourages trivial synthetic data where the teacher and student classification match, by amplifying discrepancies.
Lowest-$\Lsynth$ batch in $\ksynth$ iterations is added to $\kddata$ and oldest samples are discarded if it is full:
\[
\kddata \gets \kddata \cup \{(\mathbf{r}_b, y_b)\}_{b=1}^{B}.
\]

\paragraph{(3) Meta-update of the generator.}
The meta-update is done only in synthesis rounds (every $\tgen$ server rounds), after Step (2).
Reptile-style update is performed on the meta-parameters \cite{ReptileKD}:
\[
\genp \gets (1-\lambda)\genp + \lambda \genp',
\]
where $\lambda \in (0,1)$ is the step size. Other optimizers (e.g., momentum or Adam) can likewise be used to update $\genp$ using the same pseudo-gradient direction $(\genp' - \genp)$. This update lets $\gnrtr$ adapt to changing teacher models without catastrophic drift, preserving generalizable knowledge during federated training. Meta-learning stabilizes synthetic-data generation under asynchronous updates.

\paragraph{(4) Multi-teacher distillation.}
We distill from buffered teachers into a student $\x[s][]$, initialized as $\x[s][] \gets \x[][\supp{t}]$.
At each distillation step, we iterate over buffered teachers $k \in \mathcal{I}_t$ and draw a teacher-aware minibatch $\mathcal{B}_k$ from $\kddata$ using class-weighted sampling based on $\clp_k$.
Concretely, each element $(\mathbf{r}, y)\in\mathcal{B}_k$ is obtained by first sampling a label $y \sim \clp_k$ and then sampling $\mathbf{r}$ uniformly from $\{(\mathbf{r}',y')\in\kddata: y'=y\}$.
The student is trained to minimize KL divergence to match the teachers:
\[
\Lkd \triangleq \frac{1}{\kdbs}\sum_{k\in\mathcal{I}_t}\frac{1}{|\mathcal{B}_k|}\sum_{(\mathbf{r},y)\in\mathcal{B}_k}\hspace{-0.4em}
\text{KL}\Big(
\softm\big(\tx[k](\mathbf{r})\big)\Big\|
\softm\big(\x[s][](\mathbf{r})\big)
\Big),
\]
and the student is updated with $\x[s][] \gets \x[s][] - \eta_{\text{KD}}\nabla_{\x[s][]}\Lkd$ for $\kkd$ steps.
Finally, we define the distilled update as
\(
\upd[][\text{KD}] \;=\; \x[s][] - \x[][\supp{t}],
\) used in the aggregation rule Eq.~(\ref{eq:ours_aggr}).

\subsection{Overhead Analysis}
\label{subsec:overhead_analysis}

\nips{
\ours{} adds server-side computation from periodic generator training and multi-teacher distillation. Compared with \sync{} averaging, this overhead is non-trivial, but it is server-only and leaves communication and client computation unchanged. In AFL regimes where client computation and communication dominate wall-clock time~\cite{FedScale, zhu2021delayed}, the extra server work is typically modest in practice. 
\camera{Periodic synthesis and a small number of distillation steps further limit the cost. For a real deployment setting, server computation can also be pipelined with client progress: upon receiving an update, the server may dispatch the next local-training request using the current global state while executing synthesis, distillation, and aggregation asynchronously, which is discussed in \apdx~\ref{appx:overhead}. This trades blocking time for additional staleness, but does not remove the server's throughput limit. If arrivals exceed available capacity, practical options include reducing $\ksynth$ or $\kkd$, increasing $\tgen$, or skipping KD for selected arrivals by setting $\mybeta=0$ for those updates in Eq.~(\ref{eq:ours_aggr}).}
The choices of $\ksynth$, $\kkd$, and $\tgen$ are discussed and evaluated in \sect~\ref{sect:experiments} and \apdx~\ref{appx:overhead}, \ref{appx:overhead_server_shift}. 
\apdx~\ref{appx:overhead} reports FLOPs/runtime breakdown.
\Cref{fig:kkd_budget_main} presents a sensitivity experiment with different KD budgets. 
Since the effective server blocking time depends on the server accelerator, implementation parallelism, and communication regime, we evaluate total wall-clock time under a range of realistic server/client speed ratios. 
\sect~\ref{sect:experiments} reports the client-computation/communication dominated AFL setting where server KD is overlappable, while \apdx~\ref{appx:overhead_server_shift} evaluates wall-clock performance after adding measured blocking server overhead under several hardware configurations. 
When the server is relatively weak, the wall-clock gains may be smaller, although the accuracy improvements remain.
In many FL deployments, however, the server is substantially more capable than the participating clients, so \ours{} is particularly well suited to such settings. 
\ours{} matches \sync{} communication cost and adds only bounded memory for the teacher buffer and generator, independent of clients and rounds \mbox{(\apdx~\ref{appx:memory_overhead})}.
}

\section{Experiments and Results}
\label{sect:experiments}

We evaluate our proposed method and baselines on both image and text classification tasks under diverse system conditions. All experiments assume a large-scale FL environment with $1000$ clients, where only $10\%$ are active at any time, selected uniformly at random, reflecting the intermittent availability of edge devices. To induce data heterogeneity, we follow Dirichlet partitioning \citep{acar2021federated}, unless the dataset is naturally partitioned. High data heterogeneity, with concentration parameter \mbox{$\alpha=0.5$}, is used in the main experiments, and $\apdx$~\ref{appx:different_het_res} presents results with milder heterogeneity ($\alpha=1$).

Since our focus is on AFL, all results are presented in simulated wall-clock time, accounting for realistic computation and communication delays. To simulate system heterogeneity, we categorize clients into \emph{slow}, \emph{medium}, and \emph{fast} groups by computation and communication speeds. Each client is assigned random delay parameters based on its group. When a client participates in training, its local computation and communication durations are sampled from its random generator, ensuring stochastic time variations across rounds. This system heterogeneity across clients and rounds reflects real environments. Additional details are in \apdx~\ref{appx:exp_time_simulation}. Main results assume non-blocking server computation; Appendix~\ref{appx:overhead_server_shift} confirms the gains persist with measured overhead on practical strong-server setups. 
Results are averaged over three random seeds with shaded regions showing standard error. Hyperparameters are tuned on validation results. As asynchronous methods often fluctuate, we plot running-average performance until the peak for clarity. \apdx~\ref{appx:originaltrainingcurves} provides raw curves.

\subsection{Datasets, Models, and Baselines}

We benchmark on image and text classification tasks. \textbf{\cifart{}} and \textbf{\cifarh{}} are 10- and 100-class RGB image classification datasets, respectively, and use ResNet-18 with 11.2M parameters~\citep{resnet}. \textbf{\femnist{}} is a 62-class grayscale handwritten character classification dataset with a natural user-level partition across approximately 3500 writers~\citep{caldas2019leafbenchmarkfederatedsettings}; we use a multi-layer CNN with 421k parameters. \textbf{\news{}} is a 20-class news topic classification dataset, for which we use a pretrained T5-small transformer with 35.6M parameters~\citep{t5model}. Except for \femnist{}, we simulate 1000 clients by repeatedly sampling the dataset with 350 samples per client. For \femnist{}, we randomly select 1000 users from the natural partition as clients. Hyperparameter choices are detailed in \apdx~\ref{appx:hyperparams}.

\begin{figure}[t]
\centering
\includegraphics[width=1\textwidth]{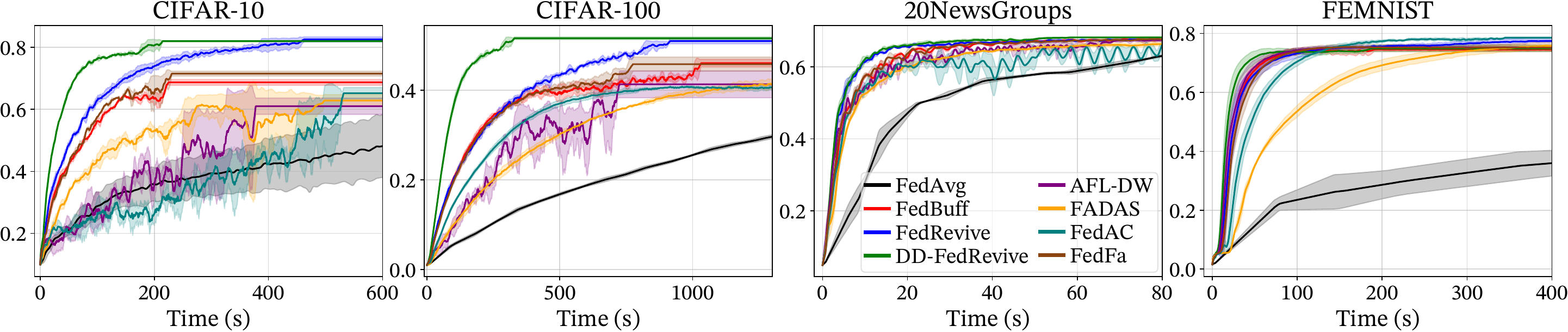}
\caption{\textbf{Test accuracy versus simulated time} on \cifart{}, \cifarh{}, \news{}, and \femnist{}.
On \cifart{}, \ours{} converges faster and reaches higher final accuracy than the baselines.
On the harder \cifarh{} task, its advantage becomes more pronounced later in training, when client models provide stronger teachers.
On \news{}, \ours{} converges faster than the baselines.
On \femnist{}, asynchronous methods show similar overall convergence due to the relative simplicity of the task, but \ours{} attains among the highest final accuracy.}
\label{fig:training_curves_alpha_05_merged}
\end{figure}

We compare \ours{} with representative baselines and ablation variants as summarized below:
\vspace{-0.5em}
\begin{itemize}[
    wide=0pt,
   topsep=1pt,
    itemsep=1pt,
    parsep=0pt,
    partopsep=0pt
]
\item \textbf{\sync:} The vanilla synchronous FL algorithm~\citep{FedAvg}. The server waits for all active clients before aggregation, which eliminates staleness, but the round time can be dominated by stragglers~\citep{xie2019asynchronous}.
\item \textbf{\fedbuff{} (Eq.~(\ref{eq:fedbuff_aggr})):} A buffered asynchronous variant of \sync{}~\cite{FedBuff}. The server accumulates incoming client updates in a buffer and aggregates once the buffer is full. When the buffer size is $1$, \fedbuff{} reduces to standard asynchronous \sync{}~\citep{xie2019asynchronous} in Eq.~(\ref{eq:afl_aggr}). Since we tune the buffer size including $1$, we report stronger \fedbuff{} as the representative asynchronous baseline.
\item \nips{\textbf{\fedfa:} An AFL baseline~\citep{fedfa}, stabilizing stale updates with a sliding window over recent updates.}
\item \nips{\textbf{\fedac:} An AFL baseline~\citep{fedac}, with buffered prospective momentum aggregation. For fair comparison of all methods, the client-side operations are kept the same as in other methods.}
\item \nips{\textbf{\fadas:} An AFL baseline~\citep{fadas}, proposing adaptive asynchronous federated optimization incorporating delayed client updates into an Adam-like server update.}
\item \textbf{\ours{} (Eq.~(\ref{eq:ours_aggr})):} Our proposed method. $\ksynth=2$, $\kkd=\tgen=10$, and $\kdbs=8$ are used in all experiments unless otherwise stated.
\item \textbf{\oursdd:} A fully equipped ablation variant, Data-Driven-\ours{}, where the server has access to a disjoint auxiliary dataset of size $16.7\%$ of the total training samples across clients, sampled from the overall client distribution. KD is performed directly on this data, i.e., \ours{} without generator-based synthesis. \oursdd{} assumes client label proportions are known here rather than estimated. This setting is included {only for ablation} due to its impractical assumptions.
\item \textbf{\asyncdw{}:} Prior asynchronous FL work~\cite{chen2019communication,9407951,9022982} down-weights stale updates to mitigate their adverse effect. To test whether \ours{} improves performance due to the KD update rather than stale-update down-weighting alone, we include AFL-Down Weighting (\asyncdw{}) as an ablation without synthesis or distillation. Concretely, we modify Eq.~(\ref{eq:ours_aggr}) to \(
\x[][\supp{t+1}] \leftarrow \x[][\supp{t}] + \lrs \bigl(1-\mybeta(\tauit)\bigr)\upd[i],
\)
while keeping all other hyperparameters and $\mybeta(\cdot)$ identical to \ours{}.
\end{itemize}

\subsection{Main Results and Ablation Experiments}
\label{sect:results}

\Cref{fig:training_curves_alpha_05_merged} and \Cref{tab:final_acc} show training curves, test accuracy and time-to-target results of main experiments. 

\paragraph{Image task results.}
For \cifart{}, \cifarh{}, and \femnist{}, the synthesis module in \ours{} uses the same lightweight generator architecture as \cite{ReptileKD}, which maps latent noise to an image via a linear projection followed by a CNN block. In all experiments, synchronous baseline \sync{} incurs long wall-clock time because the server waits for all selected clients, which limits practical convergence speed despite avoiding staleness.

\begin{wraptable}[16]{r}{0.56\textwidth}
\vspace{-1\baselineskip}
\centering
\setlength{\tabcolsep}{2pt}
\scriptsize
\caption{ \nips{
Final test accuracy and time-to-target, reported as {accuracy\% (time)}, where time is the mean simulated time to reach 85\% of \fedbuff{} peak accuracy. Per dataset, the best two values are bold and the third best is underlined, separately for accuracy and time. \ours{} provides the strongest accuracy and speed on most datasets.
}
}
\label{tab:final_acc}
\begin{tabular}{lcccc}
\toprule
\textbf{Method} & \cifart{} & \cifarh{} & \news{} & \femnist{} \\
\midrule
\sync{} &
63.18 (631.1) &
\third{45.73} (>1000) &
\textbf{68.46} (51.2) &
63.62 (>1000) \\
\fedbuff{} &
66.23 (96.7) &
44.46 (320.7) &
\third{67.93} (13.5) &
74.43 \third{(38.0)} \\
\asyncdw{} &
53.62 (---) &
37.89 (525.1) &
67.40 (14.3) &
75.07 \textbf{(35.3)} \\
\fadas{} &
59.89 (214.7) &
31.09 (---) &
67.20 (18.1) &
\third{76.90} (146.0) \\
\fedac{} &
59.71 (458.9) &
43.39 (540.0) &
67.14 (16.6) &
\textbf{78.34} (70.5) \\
\fedfa{} &
\third{70.08} \third{(90.3)} &
45.18 \third{(318.3)} &
\textbf{68.00} \third{(11.1)} &
75.43 (44.8) \\
\oursdd{} &
\textbf{79.79} \textbf{(30.2)} &
\textbf{50.01} \textbf{(99.6)} &
\third{67.93} \textbf{(7.4)} &
74.56 \textbf{(25.6)} \\
\ours{} &
\textbf{81.46} \textbf{(69.9)} &
\textbf{50.09} \textbf{(301.1)} &
67.46 \textbf{(7.8)} &
\textbf{77.42} (39.9) \\
\bottomrule
\end{tabular}
\vspace{-1.3\baselineskip}
\end{wraptable}

On \cifart{}, \ours{} reaches higher final accuracy than AFL baselines while converging faster. This gain comes from continually transferring knowledge from buffered stale client models through DFKD, so each received update contributes more effectively than parameter aggregation alone. \mbox{\oursdd{}}, the ablation variant with access to a small public dataset for distillation, exhibits the fastest convergence due to the immediate availability of real data. However, \mbox{\ours{}} surpasses \oursdd{} in final performance, as the fixed public dataset in \oursdd{} may lead to partial overfitting, while the synthetic data in \mbox{\ours{}} adaptively evolves throughout the training, yielding richer and more diverse knowledge transfer. Finally, the performance of \asyncdw{} remains considerably lower than that of \ours{}, confirming that the improvement of our method arises not primarily from down-weighting stale updates but from the proposed DFKD mechanism itself.

On the harder \cifarh{} task with $100$ classes, class-conditioned synthesis and distillation are more challenging, especially early in training when the teachers are still weak. Once client models become sufficiently accurate, the advantage of \ours{} becomes clearer and the gap with AFL baselines widens. The ablation variant \oursdd{} converges fastest due to immediate access to real server-side data and client label proportions. \ours{} achieves the highest final accuracy.
Finally, on \femnist{}, stale updates are less detrimental, likely due to simpler visual patterns, so AFL baselines exhibit similar convergence. Even in this regime, \ours{} is competitive. 

\paragraph{Text task results.}
For the text-domain task, \news{}, we adopt the T5-small architecture \citep{t5model} as the backbone and attach a lightweight classification head composed of linear layers. The head reads the final hidden state of a prepended dummy prediction token and outputs class logits. We truncate inputs to length $256$ after adding this token.
For discrete text inputs, we replace the image generator with a prompt-based synthesis mechanism. Unlike images, text is discrete and highly structured; and prior data-free distillation for language often relies on external or generated corpora~\cite{text_kd_support1, text_kd_support2, text_kd_support3}. In \ours{}, rather than training a generator, we meta-learn four soft prompt vectors inserted immediately after the prediction token, following soft prompt learning~\cite{softprompt1, softprompt2}. 
For the remaining latent prompt positions, we initialize from embeddings of short high-temperature synthetic snippets from a small LM~\cite{zhang2024tinyllama}, using them only as a linguistic warm start for optimization rather than as client data, labels, or an auxiliary training corpus.
Figure~\ref{fig:training_curves_alpha_05_merged} (third panel) shows test accuracy versus simulated time on \news{}.
\ours{} converges faster than the baselines.
We verify that the LM-generated snippets do not provide meaningful task leakage. Centralized training on only the initialization snippets achieves a best test accuracy of $48.08\%\pm0.22\%$ on \news{}, which is far below the federated performance of all methods reported in Table~\ref{tab:final_acc}. Thus, the initialization snippets only serve as a weak linguistic prior and a neutral initialization that stabilizes the synthetic optimization, rather than a competitive task dataset. In this setup, the generator parameters $\genp$ correspond to four meta-learned soft prompt vectors, while the latent variable $\noise$ denotes the remaining embedding vectors initialized from the synthetic snippets.

\begin{figure}[t]
\centering

\begin{minipage}[t]{0.37\textwidth}
\centering
\includegraphics[width=\linewidth]{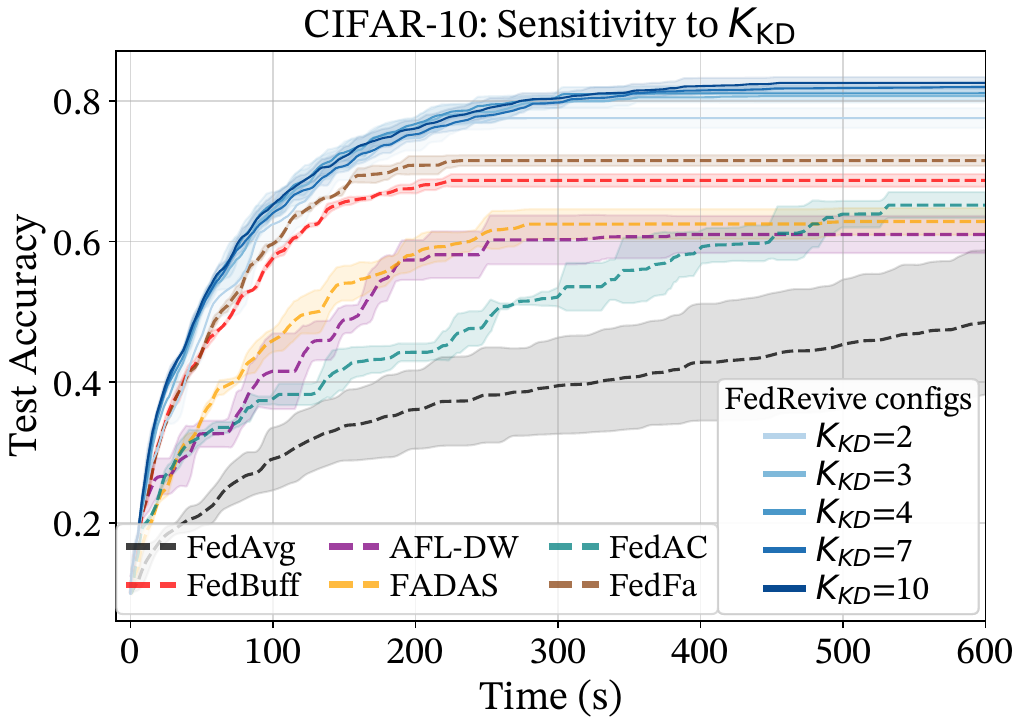}
\caption{
\nips{\textbf{Sensitivity of \ours{} to the number of KD steps $\kkd$ on \cifart{},} showing that the gain does not depend on heavy server-side computation.}
}
\label{fig:kkd_budget_main}
\end{minipage}\hfill
\begin{minipage}[t]{0.6\textwidth}
\centering
\includegraphics[width=\linewidth]{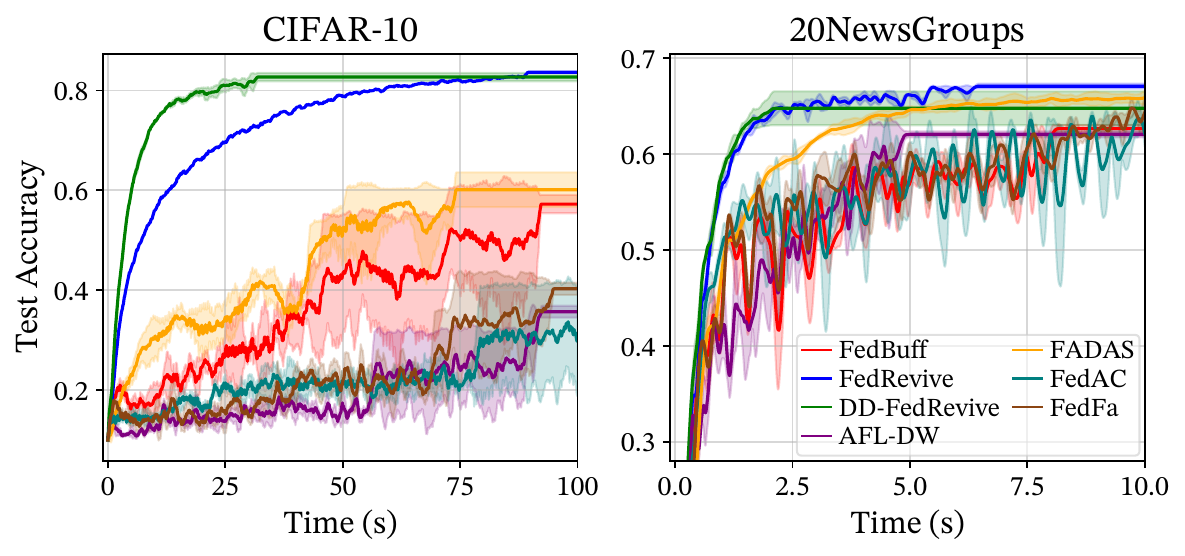}
\caption{\textbf{Robustness to a delay distribution.} Test accuracy on \cifart{} and \news{} under an alternative delay scheme with a shifted staleness distribution. Hyperparameters are tuned under the default delay model. \ours{} remains robust, while baselines degrade significantly.}
\label{fig:staleness_shift_robustness_merged}
\end{minipage}

\end{figure}

\paragraph{Robustness to shift in delay schemes.}
All main experiments use a fixed stochastic delay model for client-side computation and communication (details in \apdx~\ref{appx:exp_time_simulation}), inducing a long-tailed staleness distribution in \Cref{fig:staleness_dist}. We tune all hyperparameters under this default delay scheme.
To test robustness to system changes, we re-run the asynchronous methods on \cifart{} and \news{} under an alternative delay schedule that shifts the induced staleness distribution (see \Cref{fig:staleness_dist_shifted} in \apdx~\ref{appx:staleness_shift_robustness}) without re-tuning any hyperparameters. These runs use Dirichlet $\alpha=1$.
\Cref{fig:staleness_shift_robustness_merged} shows that the DFKD-based methods (\ours{} and \oursdd{}) remain substantially more robust to this delay shift, while asynchronous baselines (e.g., \fedbuff{} and \asyncdw{}) show slower convergence and lower final accuracy.

\nips{
\paragraph{Sensitivity to KD budget.}
Since distillation is the main server-side cost, we vary the number of KD steps $\kkd$ on \cifart{}.
As shown in \Cref{fig:kkd_budget_main}, \ours{} remains strong even with substantially fewer KD steps, indicating that its gains are not tied to a heavy server-side KD budget.
\apdx~\ref{appx:overhead_server_shift} evaluates wall-clock performance with server overhead under several hardware configurations.
}

\paragraph{Update contributions in DFKD and $\beta$ function.}
We analyze the contributions of stale client and KD-revived updates in Eq.~(\ref{eq:ours_aggr}), as well as the role of the staleness-dependent mixing schedule $\mybeta(\tau)$. For \ours{} and \oursdd{}, $\mybeta(\tau)$ controls the interpolation between the parameter-space update and the KD update: when staleness is small, the server should rely more on the raw client update, whereas for highly stale updates, the KD component should be emphasized. Results show that the KD-revived component provides strong gains, while over-reliance throughout training degrades overall performance, indicating that sustained progress still requires preserving the contribution of the original client updates across rounds. Full experimental details and plots are provided in \apdx~\ref{appx:update_ablations}. We consider several monotone scheduling families for $\mybeta$ (Figure~\ref{fig:beta_families}) and use the \mbox{$1-\mathrm{cosine}$} schedule for \ours{}, while \oursdd{} uses a constant mixing coefficient $\beta=0.5$. \nips{\ours{} is also broadly robust to the choice of $\mybeta(\tau)$, with adaptive schedules yielding better overall training behavior. See \apdx~\ref{appx:beta_function_ablation} for a detailed comparison.}

\section{Conclusion}
Asynchronous federated learning algorithms offer the scalability required for large-scale decentralized training and cross-device FL, yet suffer from staleness. Although stale updates can hinder convergence, they still contain valuable information about the clients' local data. We introduce \ours{}, a scalable asynchronous FL framework that extracts and utilizes this information through data-free knowledge distillation. By combining lightweight meta-learned synthetic data generation with multi-teacher distillation, our approach effectively mitigates staleness while preserving data privacy and system efficiency. Experiments on vision and text tasks demonstrate that \ours{} significantly improves convergence speed and final accuracy over baselines. Extending \ours{} with pretrained generative models for applicable data modalities presents a promising direction.

\camera{
\section*{Acknowledgements}
This work was partially supported by the US National Science Foundation under grants CNS-2106891 and CNS-2409138 to CJW and CCF 2045694, CNS-2112471, CPS-2111751, and ONR N00014-23-1-2149 to GJ.
}

\bibliographystyle{plainnat}
\bibliography{references}

@inproceedings{dean2012large,
  title={Large scale distributed deep networks},
  author={Dean, Jeffrey and  others},
  booktitle={Advances in Neural Information Processing Systems},
  pages={1223--1231},
  year={2012}
}

@inproceedings{cui2014exploiting,
  title={Exploiting Bounded Staleness to Speed Up Big Data Analytics},
  author={Cui, Henggang and Cipar, James and Ho, Qirong and Kim, Jin Kyu and Lee, Seunghak and Kumar, Abhimanu and Wei, Jinliang and Dai, Wei and Ganger, Gregory R and Gibbons, Phillip B and others},
  booktitle={USENIX Annual Technical Conference ({ATC})},
  pages={37--48},
  year={2014}
}

@inproceedings{ho2013more,
  title={More effective distributed ml via a stale synchronous parallel parameter server},
  author={Ho, Qirong and Cipar, James and Cui, Henggang and Lee, Seunghak and Kim, Jin Kyu and Gibbons, Phillip B and Gibson, Garth A and Ganger, Greg and Xing, Eric P},
  booktitle={Advances in Neural Information Processing Systems},
  pages={1223--1231},
  year={2013}
}

@inproceedings{zhang2016staleness,
  title={Staleness-aware async-SGD for distributed deep learning},
  author={Zhang, Wei and Gupta, Suyog and Lian, Xiangru and Liu, Ji},
  booktitle={International Joint Conference on Artificial Intelligence},
  pages={2350--2356},
  year={2016},
  organization={AAAI Press}
}

@article{dutta2018slow,
 author = {Sanghamitra Dutta and Gauri Joshi and Soumyadip Ghosh and Parijat Dube and Priya Nagpurkar},
 title = "{Slow and Stale Gradients Can Win the Race: Error-Runtime Trade-offs in Distributed SGD}",
journal = {Proceedings of the International Conference on Artificial Intelligence and Statistics (AISTATS)},
 year = {2018},
 month = apr
}

@inproceedings{gupta2016model,
 author = {Suyog Gupta and Wei Zhang and Fei Wang},
 title = {Model Accuracy and Runtime Tradeoff in Distributed Deep Learning: A Systematic Study},
 booktitle = {International Conference on Data Mining},
 year = {2016},
 pages={171-180}
}

@inproceedings{mitliagkas2016asynchrony,
  title={Asynchrony begets momentum, with an application to deep learning},
  author={Mitliagkas, Ioannis and Zhang, Ce and Hadjis, Stefan and R{\'e}, Christopher},
  booktitle={Allerton Conference on Communication, Control, and Computing},
  pages={997--1004},
  year={2016},
  organization={IEEE}
}

@inproceedings{lian2015asynchronous,
  title={Asynchronous parallel stochastic gradient for nonconvex optimization},
  author={Lian, Xiangru and Huang, Yijun and Li, Yuncheng and Liu, Ji},
  booktitle={Advances in Neural Information Processing Systems},
  pages={2737--2745},
  year={2015}
}

@inproceedings{FedScale,
author = {Lai, Fan and Dai, Yinwei and Zhu, Xiangfeng and Madhyastha, Harsha V. and Chowdhury, Mosharaf},
title = {FedScale: Benchmarking Model and System Performance of Federated Learning},
year = {2021},
isbn = {9781450387088},
publisher = {Association for Computing Machinery},
address = {New York, NY, USA},
url = {https://doi.org/10.1145/3477114.3488760},
doi = {10.1145/3477114.3488760},
booktitle = {Proceedings of the First Workshop on Systems Challenges in Reliable and Secure Federated Learning},
pages = {1–3},
numpages = {3},
location = {Virtual Event, Germany},
series = {ResilientFL '21}
}

@InProceedings{FedAvg,
  title = 	 {{Communication-Efficient Learning of Deep Networks from Decentralized Data}},
  author = 	 {McMahan, Brendan and Moore, Eider and Ramage, Daniel and Hampson, Seth and Arcas, Blaise Aguera y},
  booktitle = 	 {Proceedings of the 20th International Conference on Artificial Intelligence and Statistics},
  pages = 	 {1273--1282},
  year = 	 {2017},
  editor = 	 {Singh, Aarti and Zhu, Jerry},
  volume = 	 {54},
  series = 	 {Proceedings of Machine Learning Research},
  month = 	 {20--22 Apr},
  publisher =    {PMLR},
  url = 	 {https://proceedings.mlr.press/v54/mcmahan17a.html}
}

@INPROCEEDINGS {online_afl,
author = { Chen, Yujing and Ning, Yue and Slawski, Martin and Rangwala, Huzefa },
booktitle = { 2020 IEEE International Conference on Big Data (Big Data) },
title = {{ Asynchronous Online Federated Learning for Edge Devices with Non-IID Data }},
year = {2020},
volume = {},
ISSN = {},
pages = {15-24},
doi = {10.1109/BigData50022.2020.9378161},
url = {https://doi.ieeecomputersociety.org/10.1109/BigData50022.2020.9378161},
publisher = {IEEE Computer Society},
address = {Los Alamitos, CA, USA},
month =Dec}

@inproceedings{FedAST,
author = {Askin, Baris and Sharma, Pranay and Joe-Wong, Carlee and Joshi, Gauri},
title = {FedAST: federated asynchronous simultaneous training},
year = {2024},
publisher = {JMLR.org},
booktitle = {Proceedings of the Fortieth Conference on Uncertainty in Artificial Intelligence},
articleno = {6},
numpages = {35},
location = {Barcelona, Spain},
series = {UAI '24}
}

@InProceedings{FedBuff,
  title = 	 { Federated Learning with Buffered Asynchronous Aggregation },
  author =       {Nguyen, John and Malik, Kshitiz and Zhan, Hongyuan and Yousefpour, Ashkan and Rabbat, Mike and Malek, Mani and Huba, Dzmitry},
  booktitle = 	 {Proceedings of The 25th International Conference on Artificial Intelligence and Statistics},
  pages = 	 {3581--3607},
  year = 	 {2022},
  editor = 	 {Camps-Valls, Gustau and Ruiz, Francisco J. R. and Valera, Isabel},
  volume = 	 {151},
  series = 	 {Proceedings of Machine Learning Research},
  month = 	 {28--30 Mar},
  publisher =    {PMLR},
  url = 	 {https://proceedings.mlr.press/v151/nguyen22b.html}
}

@article{KD,
  author       = {Geoffrey E. Hinton and
                  Oriol Vinyals and
                  Jeffrey Dean},
  title        = {Distilling the Knowledge in a Neural Network},
  journal      = {CoRR},
  volume       = {abs/1503.02531},
  year         = {2015},
  url          = {http://arxiv.org/abs/1503.02531},
}

@article{ReptileKD, 
title={Up to 100x Faster Data-Free Knowledge Distillation}, 
volume={36}, 
url={https://ojs.aaai.org/index.php/AAAI/article/view/20613}, 
DOI={10.1609/aaai.v36i6.20613}, 
abstractNote={Data-free knowledge distillation (DFKD) has recently been attracting increasing attention from research communities, attributed to its capability to compress a model only using synthetic data. Despite the encouraging results achieved, state-of-the-art DFKD methods still suffer from the inefficiency of data synthesis, making the data-free training process extremely time-consuming and thus inapplicable for large-scale tasks. In this work, we introduce an efficacious scheme, termed as FastDFKD, that allows us to accelerate DFKD by a factor of orders of magnitude. At the heart of our approach is a novel strategy to reuse the shared common features in training data so as to synthesize different data instances. Unlike prior methods that optimize a set of data independently, we propose to learn a meta-synthesizer that seeks common features as the initialization for the fast data synthesis. As a result, FastDFKD achieves data synthesis within only a few steps, significantly enhancing the efficiency of data-free training. Experiments over CIFAR, NYUv2, and ImageNet demonstrate that the proposed FastDFKD achieves 10x and even 100x acceleration while preserving performances on par with state of the art. Code is available at https://github.com/zju-vipa/Fast-Datafree.}, number={6}, journal={Proceedings of the AAAI Conference on Artificial Intelligence}, author={Fang, Gongfan and Mo, Kanya and Wang, Xinchao and Song, Jie and Bei, Shitao and Zhang, Haofei and Song, Mingli}, year={2022}, month={Jun.}, pages={6597-6604} }

@article{koloskova2022sharper,
  title={Sharper convergence guarantees for asynchronous SGD for distributed and federated learning},
  author={Koloskova, Anastasiia and Stich, Sebastian U and Jaggi, Martin},
  journal={Advances in Neural Information Processing Systems},
  volume={35},
  pages={17202--17215},
  year={2022}
}

@article{rieke2020future,
  title={The future of digital health with federated learning},
  author={Rieke, Nicola and Hancox, Jonny and Li, Wenqi and Milletari, Fausto and Roth, Holger R and Albarqouni, Shadi and Bakas, Spyridon and Galtier, Mathieu N and Landman, Bennett A and Maier-Hein, Klaus and others},
  journal={NPJ digital medicine},
  volume={3},
  number={1},
  pages={119},
  year={2020},
  publisher={Nature Publishing Group UK London}
}

@article{yang2019federated,
  title={Federated machine learning: Concept and applications},
  author={Yang, Qiang and Liu, Yang and Chen, Tianjian and Tong, Yongxin},
  journal={ACM Transactions on Intelligent Systems and Technology (TIST)},
  volume={10},
  number={2},
  pages={1--19},
  year={2019},
  publisher={ACM New York, NY, USA}
}

@inproceedings{richtarik2025handling,
  title={Handling Device Heterogeneity in Federated Learning: The First Optimal Parallel SGD in the Presence of Data, Compute and Communication Heterogeneity},
  author={Richtarik, Peter},
  booktitle={Proceedings of the International Workshop on Secure and Efficient Federated Learning},
  pages={1--1},
  year={2025}
}

@article{xie2019asynchronous,
  title={Asynchronous federated optimization},
  author={Xie, Cong and Koyejo, Sanmi and Gupta, Indranil},
  journal={arXiv preprint arXiv:1903.03934},
  year={2019}
}

@article{chen2021fedsa,
  title={FedSA: A staleness-aware asynchronous federated learning algorithm with non-IID data},
  author={Chen, Ming and Mao, Bingcheng and Ma, Tianyi},
  journal={Future Generation Computer Systems},
  volume={120},
  pages={1--12},
  year={2021},
  publisher={Elsevier}
}

@article{nichol2018first,
  title={On first-order meta-learning algorithms},
  author={Nichol, Alex and Achiam, Joshua and Schulman, John},
  journal={arXiv preprint arXiv:1803.02999},
  year={2018}
}

@inproceedings{
acar2021federated,
title={Federated Learning Based on Dynamic Regularization},
author={Durmus Alp Emre Acar and Yue Zhao and Ramon Matas and Matthew Mattina and Paul Whatmough and Venkatesh Saligrama},
booktitle={International Conference on Learning Representations},
year={2021},
url={https://openreview.net/forum?id=B7v4QMR6Z9w}
}

@misc{caldas2019leafbenchmarkfederatedsettings,
      title={LEAF: A Benchmark for Federated Settings}, 
      author={Sebastian Caldas and Sai Meher Karthik Duddu and Peter Wu and Tian Li and Jakub Konečný and H. Brendan McMahan and Virginia Smith and Ameet Talwalkar},
      year={2019},
      eprint={1812.01097},
      archivePrefix={arXiv},
      primaryClass={cs.LG},
      url={https://arxiv.org/abs/1812.01097}, 
}

@misc{t5model,
      title={Exploring the Limits of Transfer Learning with a Unified Text-to-Text Transformer}, 
      author={Colin Raffel and Noam Shazeer and Adam Roberts and Katherine Lee and Sharan Narang and Michael Matena and Yanqi Zhou and Wei Li and Peter J. Liu},
      year={2023},
      eprint={1910.10683},
      archivePrefix={arXiv},
      primaryClass={cs.LG},
      url={https://arxiv.org/abs/1910.10683}, 
}

@article{XU2023100595,
title = {Asynchronous federated learning on heterogeneous devices: A survey},
journal = {Computer Science Review},
volume = {50},
pages = {100595},
year = {2023},
issn = {1574-0137},
doi = {https://doi.org/10.1016/j.cosrev.2023.100595},
url = {https://www.sciencedirect.com/science/article/pii/S157401372300062X},
author = {Chenhao Xu and Youyang Qu and Yong Xiang and Longxiang Gao}
}

@article{chen2019communication,
  title={Communication-efficient federated deep learning with layerwise asynchronous model update and temporally weighted aggregation},
  author={Chen, Yang and Sun, Xiaoyan and Jin, Yaochu},
  journal={IEEE transactions on neural networks and learning systems},
  volume={31},
  number={10},
  pages={4229--4238},
  year={2019},
  publisher={IEEE}
}

@INPROCEEDINGS{9407951,
  author={Shi, Guomei and Li, Li and Wang, Jun and Chen, Wenyan and Ye, Kejiang and Xu, ChengZhong},
  booktitle={2020 IEEE 22nd International Conference on High Performance Computing and Communications; IEEE 18th International Conference on Smart City; IEEE 6th International Conference on Data Science and Systems (HPCC/SmartCity/DSS)}, 
  title={HySync: Hybrid Federated Learning with Effective Synchronization}, 
  year={2020},
  volume={},
  number={},
  pages={628-633},
  doi={10.1109/HPCC-SmartCity-DSS50907.2020.00080}}

@ARTICLE{9022982,
  author={Lu, Xiaofeng and Liao, Yuying and Lio, Pietro and Hui, Pan},
  journal={IEEE Access}, 
  title={Privacy-Preserving Asynchronous Federated Learning Mechanism for Edge Network Computing}, 
  year={2020},
  volume={8},
  number={},
  pages={48970-48981},
  doi={10.1109/ACCESS.2020.2978082}}

@misc{text_kd_support2,
      title={A Survey on Knowledge Distillation of Large Language Models}, 
      author={Xiaohan Xu and Ming Li and Chongyang Tao and Tao Shen and Reynold Cheng and Jinyang Li and Can Xu and Dacheng Tao and Tianyi Zhou},
      year={2024},
      eprint={2402.13116},
      archivePrefix={arXiv},
      primaryClass={cs.CL},
      url={https://arxiv.org/abs/2402.13116}, 
}

@inproceedings{text_kd_support3,
author = {Tian, Yijun and Han, Yikun and Chen, Xiusi and Wang, Wei and Chawla, Nitesh V.},
title = {Beyond Answers: Transferring Reasoning Capabilities to Smaller LLMs Using Multi-Teacher Knowledge Distillation},
year = {2025},
isbn = {9798400713293},
publisher = {Association for Computing Machinery},
address = {New York, NY, USA},
url = {https://doi.org/10.1145/3701551.3703577},
doi = {10.1145/3701551.3703577},
booktitle = {Proceedings of the Eighteenth ACM International Conference on Web Search and Data Mining},
pages = {251–260},
numpages = {10},
location = {Hannover, Germany},
series = {WSDM '25}
}

@inproceedings{softprompt1,
    title = "The Power of Scale for Parameter-Efficient Prompt Tuning",
    author = "Lester, Brian  and
      Al-Rfou, Rami  and
      Constant, Noah",
    editor = "Moens, Marie-Francine  and
      Huang, Xuanjing  and
      Specia, Lucia  and
      Yih, Scott Wen-tau",
    booktitle = "Proceedings of the 2021 Conference on Empirical Methods in Natural Language Processing",
    month = nov,
    year = "2021",
    address = "Online and Punta Cana, Dominican Republic",
    publisher = "Association for Computational Linguistics",
    url = "https://aclanthology.org/2021.emnlp-main.243/",
    doi = "10.18653/v1/2021.emnlp-main.243",
    pages = "3045--3059"
}

@inproceedings{softprompt2,
    title = "{PPT}: Pre-trained Prompt Tuning for Few-shot Learning",
    author = "Gu, Yuxian  and
      Han, Xu  and
      Liu, Zhiyuan  and
      Huang, Minlie",
    editor = "Muresan, Smaranda  and
      Nakov, Preslav  and
      Villavicencio, Aline",
    booktitle = "Proceedings of the 60th Annual Meeting of the Association for Computational Linguistics (Volume 1: Long Papers)",
    month = may,
    year = "2022",
    address = "Dublin, Ireland",
    publisher = "Association for Computational Linguistics",
    url = "https://aclanthology.org/2022.acl-long.576/",
    doi = "10.18653/v1/2022.acl-long.576",
    pages = "8410--8423"
}

@misc{zhang2024tinyllama,
      title={TinyLlama: An Open-Source Small Language Model}, 
      author={Peiyuan Zhang and Guangtao Zeng and Tianduo Wang and Wei Lu},
      year={2024},
      eprint={2401.02385},
      archivePrefix={arXiv},
      primaryClass={cs.CL}
}

@inproceedings{text_kd_support1,
    title = "Performance-Guided {LLM} Knowledge Distillation for Efficient Text Classification at Scale",
    author = "Palo, Flavio Di  and
      Singhi, Prateek  and
      Fadlallah, Bilal H",
    editor = "Al-Onaizan, Yaser  and
      Bansal, Mohit  and
      Chen, Yun-Nung",
    booktitle = "Proceedings of the 2024 Conference on Empirical Methods in Natural Language Processing",
    month = nov,
    year = "2024",
    address = "Miami, Florida, USA",
    publisher = "Association for Computational Linguistics",
    url = "https://aclanthology.org/2024.emnlp-main.215/",
    doi = "10.18653/v1/2024.emnlp-main.215",
    pages = "3675--3687"
}

@inproceedings{10.1109/INFOCOM48880.2022.9796935,
author = {Luo, Bing and Xiao, Wenli and Wang, Shiqiang and Huang, Jianwei and Tassiulas, Leandros},
title = {Tackling System and Statistical Heterogeneity for Federated Learning with Adaptive Client Sampling},
year = {2022},
publisher = {IEEE Press},
url = {https://doi.org/10.1109/INFOCOM48880.2022.9796935},
doi = {10.1109/INFOCOM48880.2022.9796935},
booktitle = {IEEE INFOCOM 2022 - IEEE Conference on Computer Communications},
pages = {1739–1748},
numpages = {10},
location = {London, United Kingdom}
}

@article{DBLP:journals/corr/abs-1902-01046,
  author       = {Kallista A. Bonawitz and
                  Hubert Eichner and
                  Wolfgang Grieskamp and
                  Dzmitry Huba and
                  Alex Ingerman and
                  Vladimir Ivanov and
                  Chlo{\'{e}} Kiddon and
                  Jakub Kone{\v{c}}n{\'y} and
                  Stefano Mazzocchi and
                  H. Brendan McMahan and
                  Timon Van Overveldt and
                  David Petrou and
                  Daniel Ramage and
                  Jason Roselander},
  title        = {Towards Federated Learning at Scale: System Design},
  journal      = {CoRR},
  volume       = {abs/1902.01046},
  year         = {2019},
  url          = {http://arxiv.org/abs/1902.01046},
  eprinttype    = {arXiv},
  eprint       = {1902.01046},
  bibsource    = {dblp computer science bibliography, https://dblp.org}
}

@inproceedings{yu2023async,
  title={Async-HFL: Efficient and robust asynchronous federated learning in hierarchical IoT networks},
  author={Yu, Xiaofan and Cherkasova, Lucy and Vardhan, Harsh and Zhao, Quanling and Ekaireb, Emily and Zhang, Xiyuan and Mazumdar, Arya and Rosing, Tajana},
  booktitle={Proceedings of the 8th ACM/IEEE Conference on Internet of Things Design and Implementation},
  pages={236--248},
  year={2023}
}

@article{
mansourian2025a,
title={A Comprehensive Survey on Knowledge Distillation},
author={Amir M. Mansourian and Rozhan Ahmadi and Masoud Ghafouri and Amir Mohammad Babaei and Elaheh Badali Golezani and Zeynab yasamani ghamchi and Vida Ramezanian and Alireza Taherian and Kimia Dinashi and Amirali Miri and Shohreh Kasaei},
journal={Transactions on Machine Learning Research},
issn={2835-8856},
year={2025},
url={https://openreview.net/forum?id=3cbJzdR78B},
note={}
}

@misc{first_dfkd,
      title={Data-Free Knowledge Distillation for Deep Neural Networks}, 
      author={Raphael Gontijo Lopes and Stefano Fenu and Thad Starner},
      year={2017},
      eprint={1710.07535},
      archivePrefix={arXiv},
      primaryClass={cs.LG},
      url={https://arxiv.org/abs/1710.07535}, 
}

@inproceedings{zagoruyko2017paying,
  title={Paying More Attention to Attention: Improving the Performance of Convolutional Neural Networks via Attention Transfer},
  author={Zagoruyko, Sergey and Komodakis, Nikos},
  booktitle={International Conference on Learning Representations},
  year={2017}
}

@inproceedings{jiao-etal-2020-tinybert,
    title = "{T}iny{BERT}: Distilling {BERT} for Natural Language Understanding",
    author = "Jiao, Xiaoqi  and
      Yin, Yichun  and
      Shang, Lifeng  and
      Jiang, Xin  and
      Chen, Xiao  and
      Li, Linlin  and
      Wang, Fang  and
      Liu, Qun",
    editor = "Cohn, Trevor  and
      He, Yulan  and
      Liu, Yang",
    booktitle = "Findings of the Association for Computational Linguistics: EMNLP 2020",
    month = nov,
    year = "2020",
    address = "Online",
    publisher = "Association for Computational Linguistics",
    url = "https://aclanthology.org/2020.findings-emnlp.372/",
    doi = "10.18653/v1/2020.findings-emnlp.372",
    pages = "4163--4174"
}

@article{fl_kd_server1,
  title={Edge bias in federated learning and its solution by buffered knowledge distillation},
  author={Lee, Sangho and Yoo, Kiyoon and Kwak, Nojun},
  journal={arXiv preprint arXiv:2010.10338},
  year={2020}
}

@article{li2024federated,
  title={Federated distillation: A survey},
  author={Li, Lin and Gou, Jianping and Yu, Baosheng and Du, Lan and Yi, Zhang and Tao, Dacheng},
  journal={arXiv preprint arXiv:2404.08564},
  year={2024}
}

@misc{li2019fedmdheterogenousfederatedlearning,
      title={FedMD: Heterogenous Federated Learning via Model Distillation}, 
      author={Daliang Li and Junpu Wang},
      year={2019},
      eprint={1910.03581},
      archivePrefix={arXiv},
      primaryClass={cs.LG},
      url={https://arxiv.org/abs/1910.03581}, 
}

@article{lin2020ensemble,
  title={Ensemble distillation for robust model fusion in federated learning},
  author={Lin, Tao and Kong, Lingjing and Stich, Sebastian U and Jaggi, Martin},
  journal={Advances in neural information processing systems},
  volume={33},
  pages={2351--2363},
  year={2020}
}

@article{itahara2021distillation,
  title={Distillation-based semi-supervised federated learning for communication-efficient collaborative training with non-iid private data},
  author={Itahara, Sohei and Nishio, Takayuki and Koda, Yusuke and Morikura, Masahiro and Yamamoto, Koji},
  journal={IEEE Transactions on Mobile Computing},
  volume={22},
  number={1},
  pages={191--205},
  year={2021},
  publisher={IEEE}
}

@ARTICLE{10041112,
  author={Shang, Ertong and Liu, Hui and Yang, Zhuo and Du, Junzhao and Ge, Yiming},
  journal={IEEE Internet of Things Journal}, 
  title={FedBiKD: Federated Bidirectional Knowledge Distillation for Distracted Driving Detection}, 
  year={2023},
  volume={10},
  number={13},
  pages={11643-11654},
  doi={10.1109/JIOT.2023.3243622}}

@misc{gao2025feddtgfederateddatafreeknowledgedistillation,
      title={FedDTG: Federated Data-Free Knowledge Distillation via Three-Player Generative Adversarial Networks}, 
      author={Lingzhi Gao and Zhenyuan Zhang and Chao Wu},
      year={2025},
      eprint={2201.03169},
      archivePrefix={arXiv},
      primaryClass={cs.LG},
      url={https://arxiv.org/abs/2201.03169}, 
}

@inproceedings{zhang2022fine,
  title={Fine-tuning global model via data-free knowledge distillation for non-iid federated learning},
  author={Zhang, Lin and Shen, Li and Ding, Liang and Tao, Dacheng and Duan, Ling-Yu},
  booktitle={Proceedings of the IEEE/CVF conference on computer vision and pattern recognition},
  pages={10174--10183},
  year={2022}
}

@inproceedings{zhu2021data,
  title={Data-free knowledge distillation for heterogeneous federated learning},
  author={Zhu, Zhuangdi and Hong, Junyuan and Zhou, Jiayu},
  booktitle={International conference on machine learning},
  pages={12878--12889},
  year={2021},
  organization={PMLR}
}

@misc{lu2025correctedlatestversionmake,
      title={Corrected with the Latest Version: Make Robust Asynchronous Federated Learning Possible}, 
      author={Chaoyi Lu and Yiding Sun and Pengbo Li and Zhichuan Yang},
      year={2025},
      eprint={2504.04081},
      archivePrefix={arXiv},
      primaryClass={cs.LG},
      url={https://arxiv.org/abs/2504.04081}, 
}

@misc{qin2024knowledgedistillationfederatedlearning,
      title={Knowledge Distillation in Federated Learning: a Survey on Long Lasting Challenges and New Solutions}, 
      author={Laiqiao Qin and Tianqing Zhu and Wanlei Zhou and Philip S. Yu},
      year={2024},
      eprint={2406.10861},
      archivePrefix={arXiv},
      primaryClass={cs.LG},
      url={https://arxiv.org/abs/2406.10861}, 
}

@INPROCEEDINGS {10825769,
author = { Fan, Boyu and Jiang, Siyang and Su, Xiang and Tarkoma, Sasu and Hui, Pan },
booktitle = { 2024 IEEE International Conference on Big Data (BigData) },
title = {{ A Survey on Model-heterogeneous Federated Learning: Problems, Methods, and Prospects }},
year = {2024},
volume = {},
ISSN = {},
pages = {7725-7734},
doi = {10.1109/BigData62323.2024.10825769},
url = {https://doi.ieeecomputersociety.org/10.1109/BigData62323.2024.10825769},
publisher = {IEEE Computer Society},
address = {Los Alamitos, CA, USA},
month =Dec}

@inproceedings{
zhu2021delayed,
title={Delayed Gradient Averaging: Tolerate the Communication Latency for Federated Learning},
author={Ligeng Zhu and Hongzhou Lin and Yao Lu and Yujun Lin and Song Han},
booktitle={Advances in Neural Information Processing Systems},
editor={A. Beygelzimer and Y. Dauphin and P. Liang and J. Wortman Vaughan},
year={2021},
url={https://openreview.net/forum?id=ACFHNxVNvfk}
}

@inproceedings{resnet,
  title={Deep residual learning for image recognition},
  author={He, Kaiming and Zhang, Xiangyu and Ren, Shaoqing and Sun, Jian},
  booktitle={Proceedings of the IEEE conference on computer vision and pattern recognition},
  pages={770--778},
  year={2016}
}

@inproceedings{oort-osdi21,
  title={Oort: Efficient Federated Learning via Guided Participant Selection},
  author={Fan Lai and Xiangfeng Zhu and Harsha V. Madhyastha and Mosharaf Chowdhury},
  booktitle={USENIX Symposium on Operating Systems Design and Implementation (OSDI)},
  year={2021}
}

@article{fedfa,
  title={Fedfa: a fully asynchronous training paradigm for federated learning},
  author={Xu, Haotian and Zhang, Zhaorui and Di, Sheng and Liu, Benben and Alharthi, Khalid Ayed and Cao, Jiannong},
  journal={arXiv preprint arXiv:2404.11015},
  year={2024}
}

@article{fedac, title={Efficient Asynchronous Federated Learning with Prospective Momentum Aggregation and Fine-Grained Correction}, volume={38}, url={https://ojs.aaai.org/index.php/AAAI/article/view/29603}, DOI={10.1609/aaai.v38i15.29603}, number={15}, journal={Proceedings of the AAAI Conference on Artificial Intelligence}, author={Zang, Yu and Xue, Zhe and Ou, Shilong and Chu, Lingyang and Du, Junping and Long, Yunfei}, year={2024}, month={Mar.}, pages={16642-16650} }

@inproceedings{fadas,
  title={FADAS: Towards Federated Adaptive Asynchronous Optimization},
  author={Wang, Yujia and Wang, Shiqiang and Lu, Songtao and Chen, Jinghui},
  booktitle={International Conference on Machine Learning},
  pages={51701--51733},
  year={2024},
  organization={PMLR}
}

@inproceedings{li2026fedcd,
author = {Li, Yichen and Su, Hang and Li, Huifa and Yang, Haolin and Zhuang, Xinlin and Xue, Haochen and Wang, Haozhao and Razzak, Imran},
title = {FedCD: towards consolidated distillation for heterogeneous federated learning},
year = {2026},
isbn = {978-1-57735-906-7},
publisher = {AAAI Press},
url = {https://doi.org/10.1609/aaai.v40i28.39494},
doi = {10.1609/aaai.v40i28.39494},
numpages = {9},
series = {AAAI'26/IAAI'26/EAAI'26}
}

@INPROCEEDINGS{wang2023dafkd,
  author={Wang, Haozhao and Li, Yichen and Xu, Wenchao and Li, Ruixuan and Zhan, Yufeng and Zeng, Zhigang},
  booktitle={2023 IEEE/CVF Conference on Computer Vision and Pattern Recognition (CVPR)}, 
  title={DaFKD: Domain-aware Federated Knowledge Distillation}, 
  year={2023},
  volume={},
  number={},
  pages={20412-20421},
  doi={10.1109/CVPR52729.2023.01955}}

\appendix

\section{Extended Related Work}
\label{appndx:extended_related_work}
\paragraph{Synchronous FL.}
Since the seminal work of \citet{FedAvg}, FL has proven effective for collaboratively training models from decentralized and private data sources. Most existing FL algorithms employ \textit{synchronous} aggregation, where the server waits for all selected clients to complete their local training and then averages their model updates. However, this synchronization scheme creates a major bottleneck. In realistic environments, edge devices exhibit high variability in computation and communication capabilities (e.g., mobile phones versus embedded edge devices, Wi-Fi versus cellular connections), resulting in widely varying local training times. Consequently, in each round, the server must wait for the slowest client (the ``straggler''), leaving faster clients idle and underutilized. Several methods have been proposed to alleviate the straggler problem, such as aggregating only the updates that arrive before a predetermined time~\cite{DBLP:journals/corr/abs-1902-01046} or randomly subsampling available clients~\cite{10.1109/INFOCOM48880.2022.9796935}. However, these approaches can bias training toward inherently faster clients and still underutilize slower ones. 
The straggler issue is further exacerbated in large-scale deployments, where the growing number of clients increases the likelihood of slow update returns and communication delays~\cite{yu2023async}. As a result, synchronous aggregation inherently limits scalability and resource efficiency in federated networks. 

\paragraph{Asynchronous FL.}
To overcome the straggler bottleneck and scalability issues of synchronous aggregation, a growing body of work has explored \textit{asynchronous FL} (AFL)~\cite{xie2019asynchronous, XU2023100595, FedBuff, online_afl, richtarik2025handling, nichol2018first}. AFL builds upon a line of previous works on asynchronous distributed stochastic gradient descent (SGD) in the data-center setting \cite{dean2012large, gupta2016model, cui2014exploiting, ho2013more, lian2015asynchronous, zhang2016staleness, mitliagkas2016asynchrony, dutta2018slow}. In AFL, the server integrates client updates as soon as they arrive, either individually or in small mini-batches, without waiting for global synchronization. This design enables faster convergence by mitigating the straggler effect and improving resource utilization. However, asynchronicity introduces a new challenge: \textit{stale updates}. Because clients train on outdated versions of the global model, their updates may become obsolete by the time they reach the server. Incorporating these stale updates can inject inconsistency or even destabilize training~\cite{XU2023100595}.

To mitigate staleness, several strategies have been proposed. \citet{FedBuff} employ a buffer for incoming updates and averages them before aggregation, which smooths model update frequency and reduces variance within the server updates due to data heterogeneity despite the cost of slower server updates. Another line of work assigns lower weights to stale updates as a function of their staleness~\cite{chen2019communication,9407951,9022982, zhang2016staleness, dutta2018slow, cui2014exploiting}, reducing the influence of outdated gradients but potentially discarding valuable information from slower clients. Importantly, each client’s update encapsulates rich information about its local data distribution. In AFL, the challenge lies in how to effectively utilize this information despite staleness.
In our work, we leverage {DFKD} to extract and transfer the informative content from stale client updates to the current server model.

\paragraph{KD and DFKD.}
KD is a powerful technique for transferring learned representations of a better-performing \emph{teacher} model to a \emph{student} \cite{KD}. The standard formulation trains the student to mimic the teacher’s output logits, which encode class probabilities, inter-class similarity, and uncertainty. KD has since been widely adopted across various modalities, including vision, speech, and natural language processing, where it consistently improves generalization and model compactness \cite{mansourian2025a, zagoruyko2017paying, jiao-etal-2020-tinybert}. The vanilla KD framework, however, fundamentally requires access to a dataset on which both the teacher and the student are evaluated during distillation. In many practical cases, such as privacy-preserving learning or FL, direct access to data is unavailable, making conventional KD infeasible. This has motivated the emergence of {DFKD} \cite{first_dfkd}, which aims to perform distillation without real data. Methods typically optimize a generator or latent codes to synthesize pseudo-samples under objectives that align teacher predictions, match intermediate statistics (e.g., batch statistics), or introduce adversarial criteria to amplify teacher–student discrepancies \cite{mansourian2025a, first_dfkd, ReptileKD}. Then, these samples are utilized for distillation.

\paragraph{KD and DFKD in FL.}
KD has also been incorporated into the FL paradigm for various purposes, such as mitigating client bias and data heterogeneity, enabling architecture-heterogeneous models, and enhancing data privacy~\cite{li2024federated, li2019fedmdheterogenousfederatedlearning, lin2020ensemble, qin2024knowledgedistillationfederatedlearning, 10825769}. In FL, KD can be performed on either the \emph{server side}, where the global model distills knowledge from client models~\cite{fl_kd_server1, li2019fedmdheterogenousfederatedlearning}, or on the \emph{client side}, where clients distill knowledge from the server model or peer models~\cite{itahara2021distillation, 10041112}. 
While some of these methods assume the presence of a shared public dataset for knowledge transfer~\cite{itahara2021distillation}, some approaches adopt DFKD to remove this dependency~\cite{gao2025feddtgfederateddatafreeknowledgedistillation, zhang2022fine, zhu2021data}. 
However, existing works applying KD or DFKD have been primarily designed for \emph{synchronous} FL, and the use of KD in \emph{asynchronous} FL remains largely unexplored. The few existing efforts~\cite{lu2025correctedlatestversionmake} rely on the unrealistic assumption of public data availability. In this work, we leverage DFKD in the asynchronous FL setting to enhance both scalability and convergence stability.


\section{Deriving Client Class-Proportion Proxies}
\label{appx:estimated_client_distribution}

\ours{} uses proxy per-client class-proportion vectors to (i) construct client-conditioned synthesis targets and (ii) perform class-aware multi-teacher distillation.
To avoid requiring access to private client label histograms, we derive a coarse per-client class-proportion proxy on the server using only the first few \emph{regular} model updates (no extra client computation, metadata, or communication).

\paragraph{Deriving class-proportion proxies from early uploads.}
Let $\mathbf{y}_i^{(u)}$ denote the $u$-th model upload received from client $i$ (ordered by server reception).
We find proxy class proportions from the first $T_{\text{est}}=2$ uploads.
For each upload $\mathbf{y}_i^{(u)}$, the server probes the model with a batch of random inputs $\mathbf{r}$ and computes
\[
\widetilde{\boldsymbol{\pi}}_i^{(u)}
=
(1/|\mathbf{r}|)\,{\textstyle\sum_{r\in\mathbf{r}}}
\softm\!\left(\mathbf{y}_i^{(u)}(r)/\text{T}_{\text{probe}}\right),
\qquad u=1,\dots,T_{\text{est}},
\]
where $\softm(\cdot)$ is the $\mathrm{softmax}$ function.
For vision models, $r$ is i.i.d.\ Gaussian noise with the same input shape as the model.
For language models, $r$ is a random token sequence with the same length as the model input (tokens sampled uniformly from the vocabulary).
The final proxy is the average across early uploads,
\[
\hat{\boldsymbol{\pi}}_i
=
(1/T_{\text{est}})\,{\textstyle\sum_{u=1}^{T_{\text{est}}}}
\widetilde{\boldsymbol{\pi}}_i^{(u)},
\]
and is fixed for the remainder of training.
We use random-input probing as a weak-signal heuristic: local training generally induces class-dependent logit biases.
As a result, even under out-of-distribution probes (Gaussian noise for vision, random tokens for text), the model's predictive distribution is not arbitrary; it is typically closer to uniform when the client's label histogram is near-uniform, and becomes increasingly skewed toward the majority classes as the client's data become more imbalanced.
We do not claim this yields exact histograms, but only a coarse ranking of dominant classes, which we find sufficient to weight buffered teachers.

This estimation is performed online as uploads arrive and requires only server-side forward passes.  We apply temperature scaling in the probing softmax.
We use $\text{T}_{\text{probe}}=0.8$ on \cifart{}, $\text{T}_{\text{probe}}=1$ on \cifarh{}, $\text{T}_{\text{probe}}=3$ on \femnist{}, and $\text{T}_{\text{probe}}=1.6$ on \news{}.

\paragraph{Proxy quality.}
For analysis only, we compare the proxy $\hat{\boldsymbol{\pi}}_i$ against the ground-truth client label distribution $\boldsymbol{\pi}_i$ via
$D_{\mathrm{KL}}(\boldsymbol{\pi}_i\|\hat{\boldsymbol{\pi}}_i)$.
For each seed, we compute the mean KL across all clients ($N=1000$), then report mean $\pm$ std across seeds in Table~\ref{tab:client_label_kl}.

\begin{table}[h]
\centering
\caption{
KL divergence between true client label proportions and the server-derived proxy proportions, reported as mean $\pm$ std across seeds.
We report the average client KL after the second received update.
}
\label{tab:client_label_kl}
\begin{tabular}{lc}
\toprule
\textbf{Dataset} & \textbf{Average $D_{\mathrm{KL}}(\boldsymbol{\pi}_i\|\hat{\boldsymbol{\pi}}_i)$} \\
\midrule
\cifart{}  & 0.4240 $\pm$ 0.0103 \\
\cifarh{}  & 0.5539 $\pm$ 0.0174 \\
\news{}    & 0.5889 $\pm$ 0.0312 \\
\femnist{} & 0.4561 $\pm$ 0.0554 \\
\bottomrule
\end{tabular}
\end{table}

\begin{figure}[h]
  \centering
  \includegraphics[width=\linewidth]{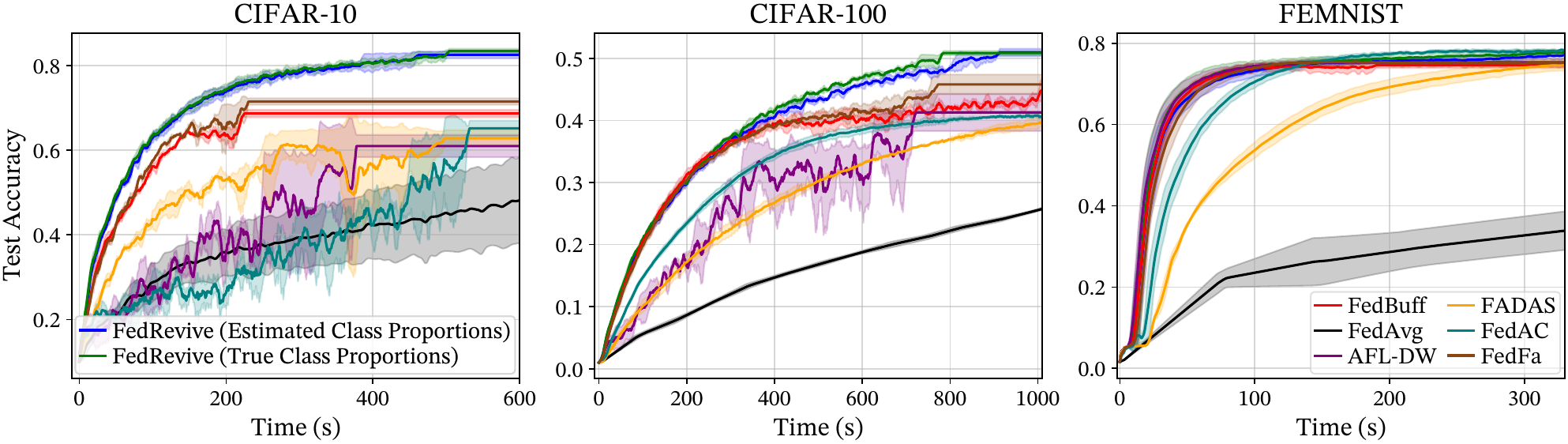}
  \vspace{-0.5em}
  \caption{\textbf{FedRevive with estimated client label proportions on \cifart{}, \cifarh{}, and \femnist{}.} Replacing ground-truth client label proportions with our probing-based proxies yields similar training dynamics and final performance across datasets.}
  \label{fig:estimated_client_dist_training_curves}
\end{figure}

\paragraph{Sensitivity to label-proportion estimation errors.}
While the probing-based proxy is only a coarse heuristic, it is designed to recover \emph{relative} class dominance rather than exact client histograms.
To evaluate whether estimation errors affect FedRevive, we run an ablation where the class-aware teacher weighting uses either the estimated proportions or the ground-truth client label proportions (just for an ablation purpose).
As shown in Fig.~\ref{fig:estimated_client_dist_training_curves}, using estimated proportions matches the training dynamics and final accuracy across \cifart{}, \cifarh{}, and \femnist{}, indicating that FedRevive is not overly sensitive to estimation noise and mainly benefits from a rough identification of dominant classes.

\nips{
\paragraph{Dominant-class recovery.}
Because the proxy is used only to provide a coarse dominant-class cue, exact histogram matching is not the most relevant metric. We therefore additionally measure whether the proxy correctly identifies the client's dominant class (top-1) and whether the true dominant class is contained in the proxy's top-2 predicted classes. Table~\ref{tab:client_label_topk} reports these accuracies across clients. These metrics better reflect the role of the proxy in class-aware teacher weighting than KL alone.

\begin{table}[h]
\centering
\caption{
\nips{Top-1 and top-2 dominant-class recovery of the server-derived proxy proportions, reported as mean $\pm$ std across seeds.
Top-1 accuracy checks whether $\arg\max_j \hat{\pi}_{i,j}$ matches the true dominant class of client $i$; top-2 accuracy checks whether the true dominant class appears in the proxy's top-2 classes.}
}
\label{tab:client_label_topk}
\begin{tabular}{lccc}
\toprule
\textbf{Dataset} & \textbf{Nb. of classes} & \textbf{Top-1 acc.} & \textbf{Top-2 acc.} \\
\midrule
\cifart{}  & 10  & $0.456 \pm 0.022$ & $0.673 \pm 0.032$ \\
\cifarh{}  & 100 & $0.135 \pm 0.001$ & $0.219 \pm 0.010$ \\
\news{}    & 20  & $0.198 \pm 0.015$ & $0.276 \pm 0.062$ \\
\femnist{} & 62  & $0.175 \pm 0.045$ & $0.338 \pm 0.014$ \\
\bottomrule
\end{tabular}
\end{table}
}

\camera{
\paragraph{Effect of label skew on dominant-class recovery.}
To isolate the effect of label skew, we evaluate the proxy on $1000$ clients under Dirichlet partitions with varying concentration $\alpha$. For each setting, clients perform one local training round from a common randomly initialized model, and we estimate each proxy from the resulting upload without server training. We use the same top-1 and top-2 recovery metrics as above. This controlled study differs from Table~\ref{tab:client_label_topk}, which evaluates proxies from early received uploads during full asynchronous training. The numerical results are therefore not directly comparable.

\begin{table}[t]
\centering
\small
\setlength{\tabcolsep}{5pt}
\caption{\camera{Dominant-class recovery across Dirichlet concentration levels after one local training round from random initialization. Entries are recovery rates across $1000$ clients; smaller $\alpha$ indicates stronger label skew.}}
\label{tab:proxy_label_skew}
\begin{tabular}{llrrrrrr}
\toprule
\textbf{Dataset} & \textbf{Metric}
& \multicolumn{6}{c}{Dirichlet concentration $\alpha$} \\
\cmidrule(lr){3-8}
& & 0.05 & 0.10 & 0.30 & 0.50 & 1.00 & 5.00 \\
\midrule
\cifart{} & Top-1 & 0.930 & 0.899 & 0.822 & 0.741 & 0.691 & 0.437 \\
          & Top-2 & 0.995 & 0.989 & 0.960 & 0.915 & 0.872 & 0.611 \\
\midrule
\cifarh{} & Top-1 & 0.708 & 0.671 & 0.528 & 0.472 & 0.382 & 0.289 \\
          & Top-2 & 0.876 & 0.847 & 0.679 & 0.626 & 0.551 & 0.468 \\
\bottomrule
\end{tabular}
\end{table}

As shown in Table~\ref{tab:proxy_label_skew}, both recovery metrics improve as label skew increases, consistent with stronger class-dependent logit biases after local training. Near-uniform client distributions provide weaker dominant-class cues. These results indicate that stronger label heterogeneity improves, rather than degrades, the proxy's ability to identify dominant classes in the evaluated settings.
}

\section{FLOPs Analysis and Server Overhead Discussion}
\label{appx:overhead}

We quantify the additional computation introduced by \ours{} on the server by reporting FLOPs per processed sample for the dominant components of the pipeline: (i) client local training, (ii) server generator training and pseudo-sample synthesis, and (iii) server-side multi-teacher distillation.
Table~\ref{tab:server_flops_breakdown} compares these costs under our default hyperparameters.

\begin{table}[h]
\centering
\caption{\textbf{Compute overhead breakdown (FLOPs) per sample for \cifart{} experiments.} FLOPs (G) for dominant components. $\kdbs$ denotes the teacher buffer size.
\\
{\textsuperscript{\textdagger}Rows report total compute except the last row, which reports effective blocking compute with concurrent implementation (total FLOPs unchanged).}
}
\label{tab:server_flops_breakdown}
\vspace{1mm}
\begin{tabular}{l c c}
\hline
Component & FLOPs (G) per iter\textsuperscript{\textdagger} & Total across iters \\
\hline
Client local training (25 iters) & ${1.65}$ & $\mathbf{41.25}$ \\
\hline
Generator train\&synth. (every $\tgen$ rounds, amortized) &
$0.14 + 0.11\,c$ &
{$\mathbf{0.28 +  0.22\,c}$} \\
Distillation (sequential implementation, 10 iters) &
$1.65 + 0.55\,c$ &
$16.5+5.5\,c$ \\
Distillation (effective cost w/ concurrent implement.) &
$2.2$ &
{$\mathbf{22}$} \\
\hline
\end{tabular}
\end{table}

First, generator training and synthesis is performed for only $\ksynth=2$ iterations and only once every $\tgen=10$ server updates, so its average cost is amortized by a factor of $\tgen=10$ (second row block in Table~\ref{tab:server_flops_breakdown}).
Second, distillation is run for $\kkd$=10 iterations, and its cost decomposes into student updates plus teacher forward passes.
Since the server maintains a buffer of teachers, forward passes for all buffered teachers except the most recent arrival can be executed asynchronously while waiting for the next client upload; thus, the effective blocking compute can be close to the student-update cost (last row in Table~\ref{tab:server_flops_breakdown}).

Third, server computation can be pipelined with client progress in asynchronous implementations. In a fully pipelined implementation, upon receiving a client update, the server may immediately dispatch the next local-training request using the current global state, and then perform synthesis, distillation, and hybrid aggregation in parallel before subsequent dispatches. This trades blocking server time for a small amount of additional staleness, which is already intrinsic to AFL. Teacher forward passes for buffered models and periodic synthesis can also be executed while the server is waiting for subsequent client uploads. We therefore model server-side KD as overlapped/non-blocking in the main simulator, but do not rely on this assumption being exact in every deployment. To account for settings where server computation is partially or fully blocking, \apdx~\ref{appx:overhead_server_shift} reports a wall-clock evaluation with measured server overhead under several hardware configurations.

Fourth, moderate server-side overhead is typically less critical than client-side computation and straggling, since modern servers are substantially more compute-capable than edge devices~\citep{DBLP:journals/corr/abs-1902-01046}.
Moreover, in many practical FL deployments, especially wireless settings, communication latency can dominate end-to-end runtime~\citep{FedAvg,zhu2021delayed}, so additional server compute often has limited impact on overall wall-clock training time.

Finally, we did not tune $\ksynth,\kkd,\tgen$ to minimize FLOPs, since our focus is better convergence performance.
In deployments where the server budget is tighter, $\ksynth$ and $\kkd$ can be reduced and $\tgen$ increased to lower the overhead.

\section{Details on Time Simulation in Experiments}
\label{appx:exp_time_simulation}
\paragraph{Delay simulation model.}
Our simulator assigns each client $i$ three independent delay generators: local computation delay $d_i^{\text{train}}$ (local training), downlink delay $d_i^{\text{down}}$ (model download), and uplink delay $d_i^{\text{up}}$ (update upload).
A client’s realized end-to-end round-trip time is the sum of these components, and the staleness $\tau$ is induced by how many server updates occur while an update is in transit and being computed. Since the servers are generally equipped with much faster hardware compared to client computation and communication delays, we model server-side operation time as non-blocking or overlapped in our simulation, supported by measured runtimes discussed in detail next.
To model persistent heterogeneity, we first randomly sample a \emph{client-specific mean} for each delay component, and keep it fixed for that client throughout training. The instantaneous delay values then vary around this mean according to the chosen distribution.
The three components are sampled independently per client, so a client may be slow in computation but fast in communication, or vice versa.

\paragraph{Server-side runtime.}
We benchmark the wall-clock cost of the main computation blocks in our CIFAR-10 setup with batch size 32 under three representative configurations.
On a 16-core CPU, a single client local-training run takes $4.874$ s in total.
Even when clients are equipped with a Tesla T4, local training still takes $0.657$ s per run.
In contrast, when the server is equipped with a modern accelerator (H100), the amortized synthesis runtime is $0.038$ s per server update (with one generator update every 10 rounds), while distillation takes $0.110$ s with a sequential implementation and $0.0569$ s with an efficient non-blocking implementation.
Furthermore, communication is typically the dominant bottleneck in cross-device FL~\cite{FedScale, zhu2021delayed}. Therefore, we model comparatively small server-side operation time as non-blocking or overlapped in our simulation.
Moreover, the server-side pipeline can be arranged to avoid introducing any additional wall-clock delay. The server can dispatch the next local-training request immediately upon receiving an update and run synthesis, distillation, and hybrid aggregation asynchronously afterward, which hides most server compute behind client-side latency at the cost of only a small additional staleness.

\paragraph{Mixture over client speeds.}
For each delay component, the client-specific mean is drawn from a discrete mixture distribution specified by a set of pairs $\{(p_k,\mu_k)\}_k$, where $\sum_k p_k=1$.
Concretely, with probability $p_k$ the client is assigned mean delay $\mu_k$ for that component.
This mixture produces multiple ``speed tiers'' and yields a long-tailed staleness distribution when combined with asynchronous execution.

\paragraph{Delay distributions and parameters.}
In our default configuration, the three delay components use different distributions:
(i) \emph{Local training delay} is exponential with a client-specific mean drawn from $\{(0.25,1.0),(0.5,1.3),(0.25,1.6)\}$.
(ii) \emph{Download delay} is a constant value $0.1$ for all clients.
(iii) \emph{Upload delay} is uniform around a client-specific mean drawn from $\{(0.5,0.15),(0.5,0.25)\}$, using a half-width of $0.02$ (i.e., sampled from $[\max(0,\mu-0.02),\,\mu+0.02]$).
All delay quantities are in the simulator's time units, and the above settings are used for all main experiments unless stated otherwise.

\begin{figure}[h]
\centering
\includegraphics[width=0.5\textwidth]{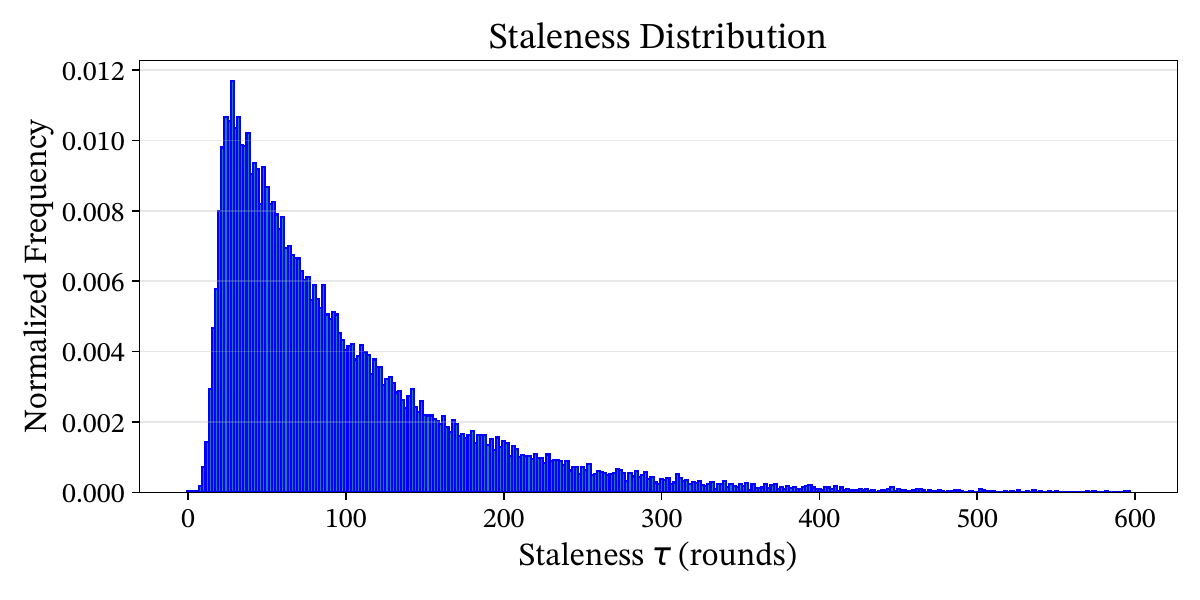}
\caption{Distribution of update staleness across server rounds for a fully asynchronous setup. The distribution aligns with real-world AFL systems reported by \citet{FedBuff}.}
\label{fig:staleness_dist}
\end{figure}

We plot the distribution of update staleness in server rounds from a sample experiment for a fully asynchronous algorithm in Figure~\ref{fig:staleness_dist}. It is shown in the figure that the staleness distribution is long-tailed due to slow clients. The figure further validates the simulation’s alignment with real practice, matching the staleness distribution observed in real-world experiments with millions of clients \citep{FedBuff}. 

\subsection{Wall-Clock Evaluation with Server Overhead under Different Hardware Configurations}
\label{appx:overhead_server_shift}

\begin{figure}[h]
\centering
\includegraphics[width=0.95\textwidth]{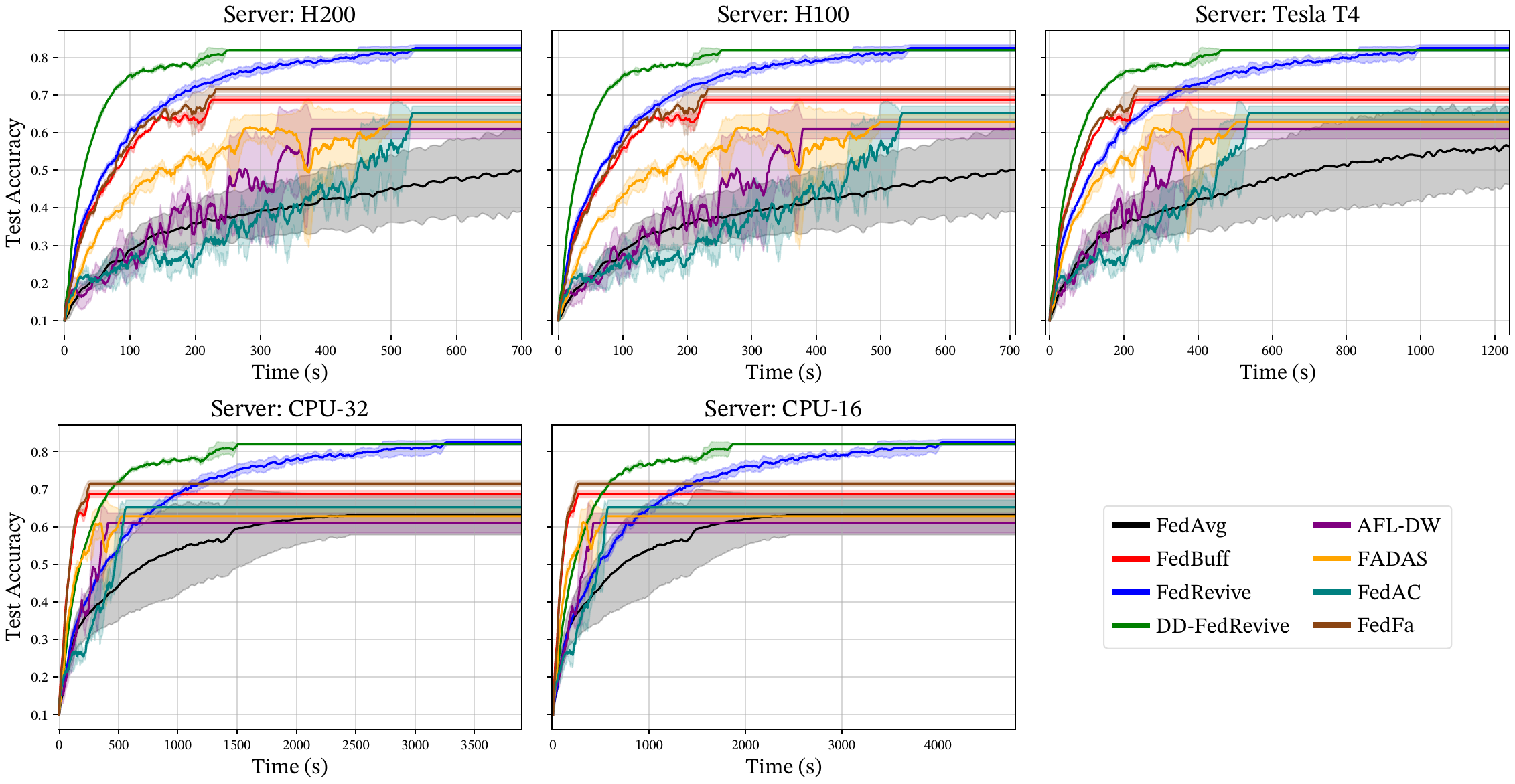}
\caption{\nips{Wall-clock evaluation with server overhead on CIFAR-10 under different hardware assumptions.
We adjust the CIFAR-10 timelines by adding measured and converted per-aggregation server runtimes for several
server configurations (H200, H100, Tesla T4, CPU-32, and CPU-16), while leaving the
underlying update sequence and training dynamics unchanged.  With strong accelerators, the conclusions are essentially unchanged;
with weaker servers, \ours{} and \oursdd{} slow down in wall-clock time but still retain a clear
accuracy advantage over the asynchronous baselines.}}
\label{fig:server_shift_panel}
\end{figure}

Experiments in \sect~\ref{sect:experiments} treat server-side KD work as overlapped or non-blocking. To evaluate the effect of explicit server blocking time, we adjust the CIFAR-10 wall-clock timelines post hoc with measured per-aggregation server runtimes under several server configurations. For an evaluation record at server round $N$ and simulated timestamp $T$, we compute the server-overhead-adjusted timestamp as
\[
T' = T + N d_{\mathrm{sim}},
\]
where $d_{\mathrm{sim}}$ is the measured per-aggregation server runtime converted into simulator time units. For this conversion, we assume edge-device clients that are slower than the servers. We take the communication-to-computation time ratio of mobile clients from~\cite{oort-osdi21}. Only \ours{} and DD-\ours{} are adjusted in this evaluation; the baselines are left unchanged because they do not perform server-side KD.
This preserves event order, staleness, and all training dynamics, and only asks what wall-clock curve the same run would report if each aggregation incurred an additional blocking server delay. Results can be seen in \Cref{fig:server_shift_panel}.

The blocking server time depends on the server accelerator, implementation parallelism, client hardware, and communication regime. Therefore, we do not insert one fixed server delay into the main simulator, which would tie the wall-clock results to one deployment choice. Instead, the main simulator models the client-side delays that cause asynchrony and staleness. We account for server-side cost separately through measured runtimes, FLOPs, and \Cref{fig:server_shift_panel}.

\subsection{Modified Delay Schedule and Shifted Staleness Distribution}
\label{appx:staleness_shift_robustness}

\begin{figure}[h]
\centering
\includegraphics[width=0.5\textwidth]{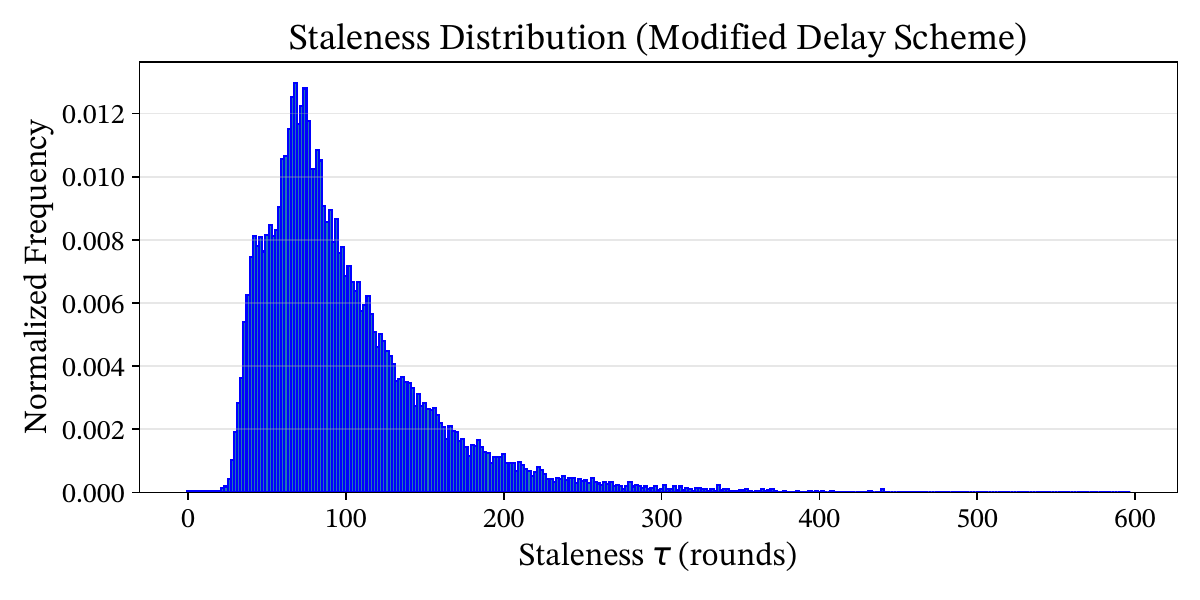}
\caption{Staleness distribution induced by the modified delay schedule used in the experiment for robustness to change in delay scheme in \sect~\ref{sect:results}. For reference, the default distribution is shown in \Cref{fig:staleness_dist}.}
\label{fig:staleness_dist_shifted}
\end{figure}

In the experiment for robustness to change in delay scheme in \sect~\ref{sect:results},
to evaluate robustness to system changes, we modify the parameters of the computation and communication delay distributions in the simulator, which yields a different induced staleness distribution (\Cref{fig:staleness_dist_shifted}) than the default setting used throughout the main experiments (\Cref{fig:staleness_dist}). In this robustness evaluation, we keep exactly the same hyperparameters for every method as tuned under the default delay scheme, and only change the delay sampling mechanism.

\paragraph{Modified delay parameters.}
Under the modified delay schedule, we use: (i) an exponential local training delay whose client-specific mean is set to $0.1$, (ii) a constant download delay with mean $0.02$ for all clients, (iii) a uniform upload delay with client-specific mean sampled from $\{(0.5,0.05),(0.5,0.10)\}$ and half-width $0.01$ (i.e., sampled from $[\max(0,\mu-0.01),\,\mu+0.01]$).
All other aspects of the simulator remain unchanged.

\section{Results with a Milder Heterogeneity Level}
\label{appx:different_het_res}

\begin{figure}[H]
  \centering
  \includegraphics[width=\linewidth]{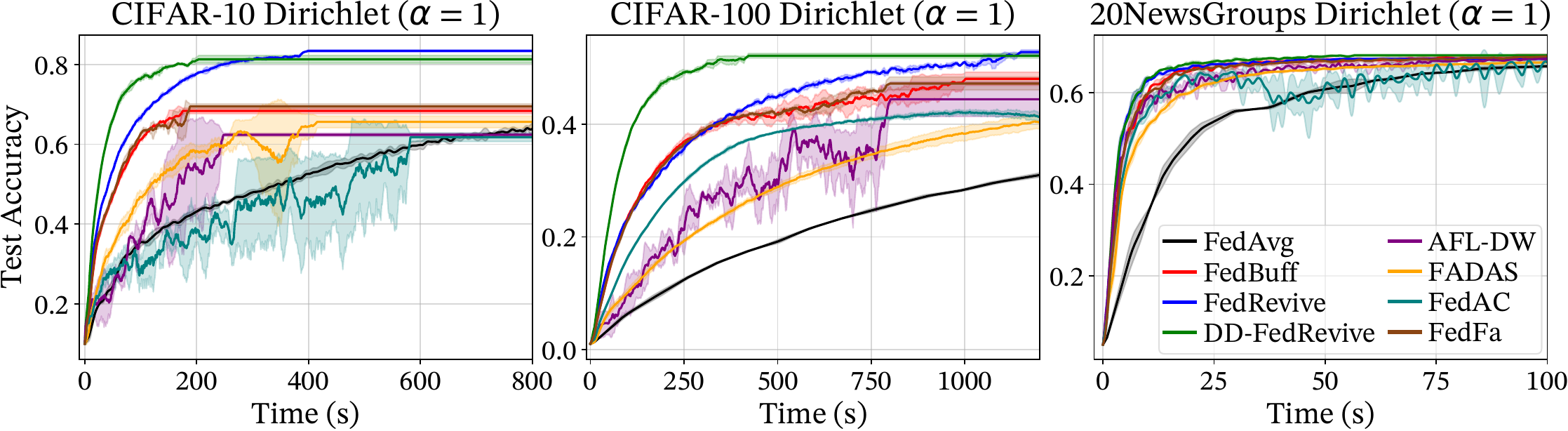}
  \caption{Test accuracy over simulated time on \cifart{}, \cifarh{}, and \news{} under milder heterogeneity ($\alpha=1$). \ours{} converges faster than \fedbuff{} and achieves higher final accuracy across datasets.}
  \label{fig:alpha1_three}
\end{figure}

We provide additional experimental results under different levels of data heterogeneity. For the \cifart{}, \cifarh{}, and \news{} datasets, we show the test accuracy curves when client data is distributed with milder non-IID partition ($\alpha=1$).
See Figure~\ref{fig:alpha1_three} for the results. Note that the \femnist{} dataset is naturally partitioned by real users and thus does not allow controlling its heterogeneity. Hence, it is excluded from this analysis. All hyperparameters except data heterogeneity are kept the same as those of main experiments.

\section{Raw Training Curves without Running Average Processing}
\label{appx:originaltrainingcurves}

\begin{figure}[h]
\centering

\begin{subfigure}{0.43\textwidth}
  \centering
  \includegraphics[width=\linewidth]{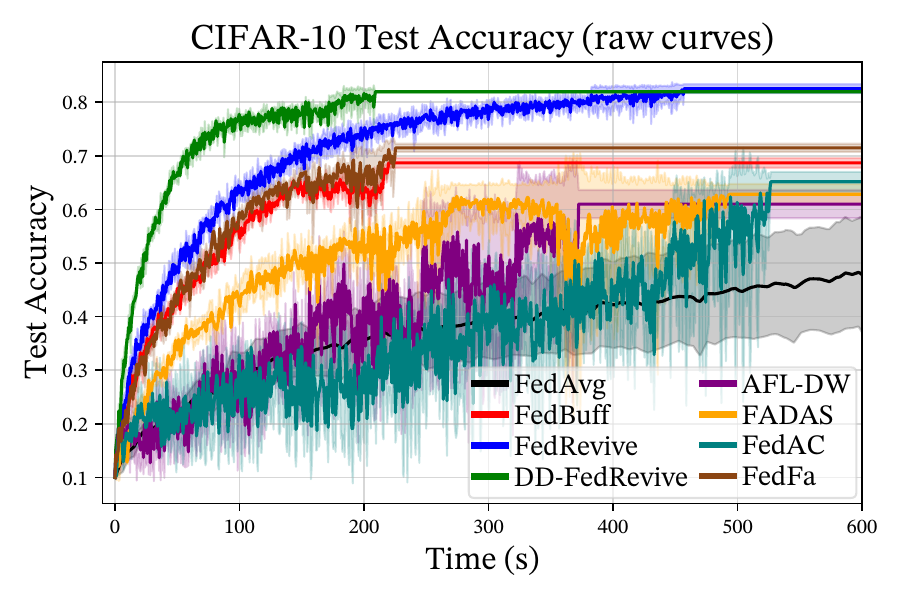}
  \vspace{-2em}
  \caption{\cifart{} raw test accuracy curves.}
  \label{fig:cifar10_raw}
\end{subfigure}\hfill
\begin{subfigure}{0.43\textwidth}
  \centering
  \includegraphics[width=\linewidth]{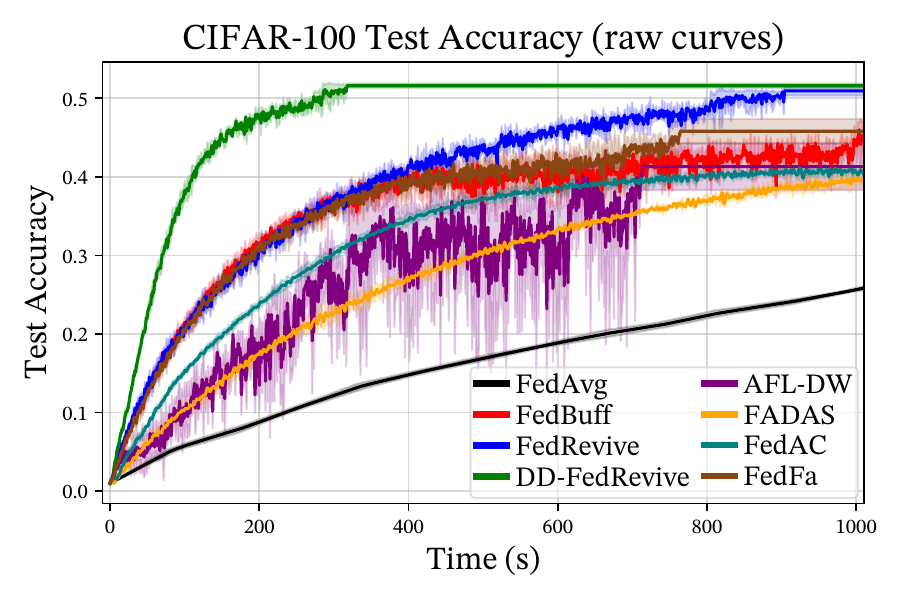}
  \vspace{-2em}
  \caption{\cifarh{} raw test accuracy curves.}
  \label{fig:cifar100_raw}
\end{subfigure}

\vspace{0.5em}

\begin{subfigure}{0.43\textwidth}
  \centering
  \includegraphics[width=\linewidth]{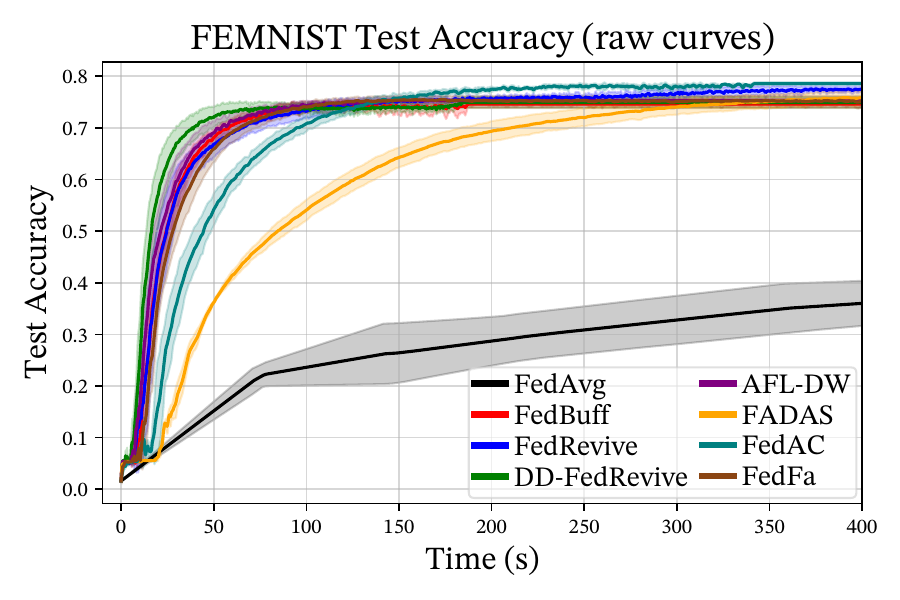}
  \vspace{-2em}
  \caption{\femnist{} raw accuracy curves.}
  \label{fig:femnist_raw}
\end{subfigure}\hfill
\begin{subfigure}{0.43\textwidth}
  \centering
  \includegraphics[width=\linewidth]{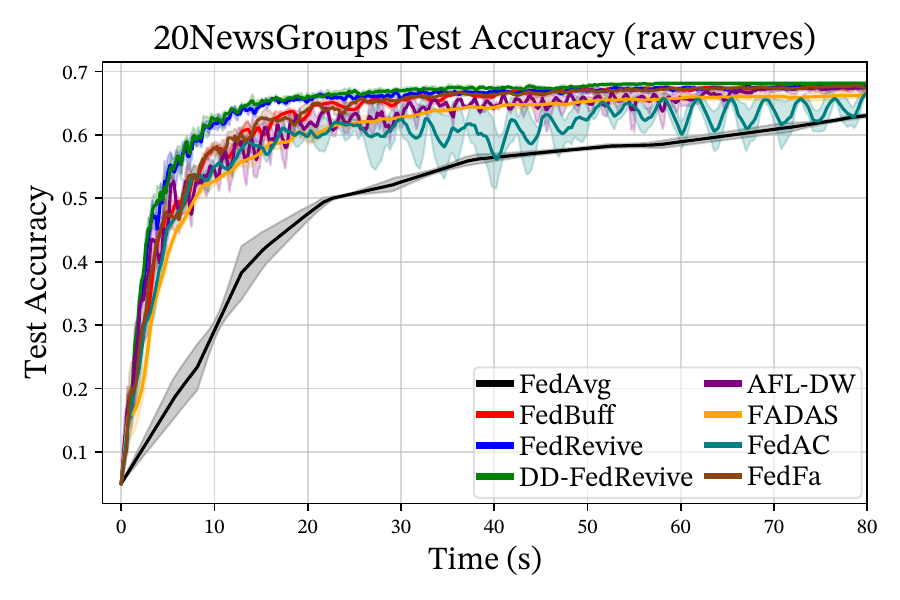}
  \vspace{-2em}
  \caption{\news{} raw accuracy curves.}
  \label{fig:news_raw}
\end{subfigure}

\caption{Raw accuracy curves across datasets ($\alpha=0.5$).}
\label{fig:raw_2x2}
\end{figure}

As described in the main text, asynchronous algorithms often display fluctuations in performance due to delayed updates and highly heterogeneous model aggregation. To provide a clearer comparison, the main figures report the moving-averaged test accuracy over time. Here, we present the original unprocessed curves showing the raw test accuracy trajectories. Figure~\ref{fig:raw_2x2} illustrates these curves.

\section{Update Ablations for DFKD and the Role of $\mybeta(\tau)$}
\label{appx:update_ablations}

This section provides a more detailed analysis of the update components in Eq.~(\ref{eq:ours_aggr}) and of the role of the staleness-dependent mixing schedule $\mybeta(\tau)$.
We study these effects on \cifart{} by separating the contribution of the stale client update and the KD-revived update produced by $\kdrevive(\cdot)$.

At each server round, we consider three update variants:
(i) only the stale client update $\upd[i][\supp{t-\tauit}]$,
(ii) only the KD-revived update $\upd[i][\text{KD}]$, and
(iii) their combined update as in Eq.~(\ref{eq:ours_aggr}).
We evaluate these variants under three mixing regimes:
(a) the adaptive staleness-dependent schedule used in the main experiments,
(b) a fixed schedule with $\mybeta(\tau)=0.75$, and
(c) a fixed schedule with $\mybeta(\tau)=1$.

\begin{figure}[t]
\centering
\includegraphics[width=0.65\textwidth]{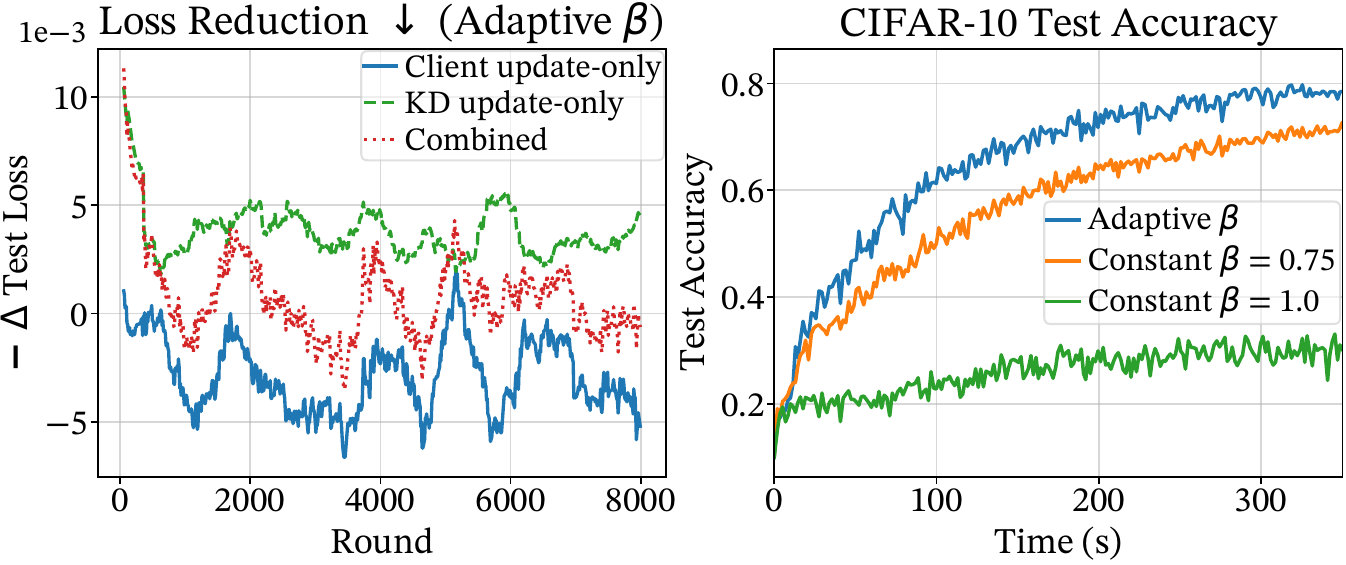}
\caption{\textbf{Update contributions and the role of $\mybeta(\tau)$ on \cifart{}.}
\textbf{Left:} Per-round effect of applying only the stale client update, only the KD-revived update, or their combination under the adaptive $\mybeta(\tau)$ schedule.
\textbf{Right:} Test accuracy vs.\ simulated time under adaptive and constant ($0.75$ and $1$) $\mybeta(\tau)$.
Constant schedules underweight the client update and degrade performance.}
\label{fig:update_contrib_dfkd}
\end{figure}

\Cref{fig:update_contrib_dfkd} summarizes the main findings.
The left panel shows that the KD-revived update yields large per-round improvements, indicating that the DFKD component is a major source of the gains of \ours{}.
At the same time, these per-update gains do not imply that training should rely on KD alone.
The right panel shows that fixed high values of $\mybeta$ lead to worse end-to-end training trajectories than the adaptive schedule.
This indicates that sustained progress still requires preserving the contribution of the original client updates across rounds, while using KD more aggressively only when staleness is high.

\begin{figure}[t]
\centering

\begin{subfigure}{0.48\textwidth}
  \centering
  \includegraphics[width=\textwidth]{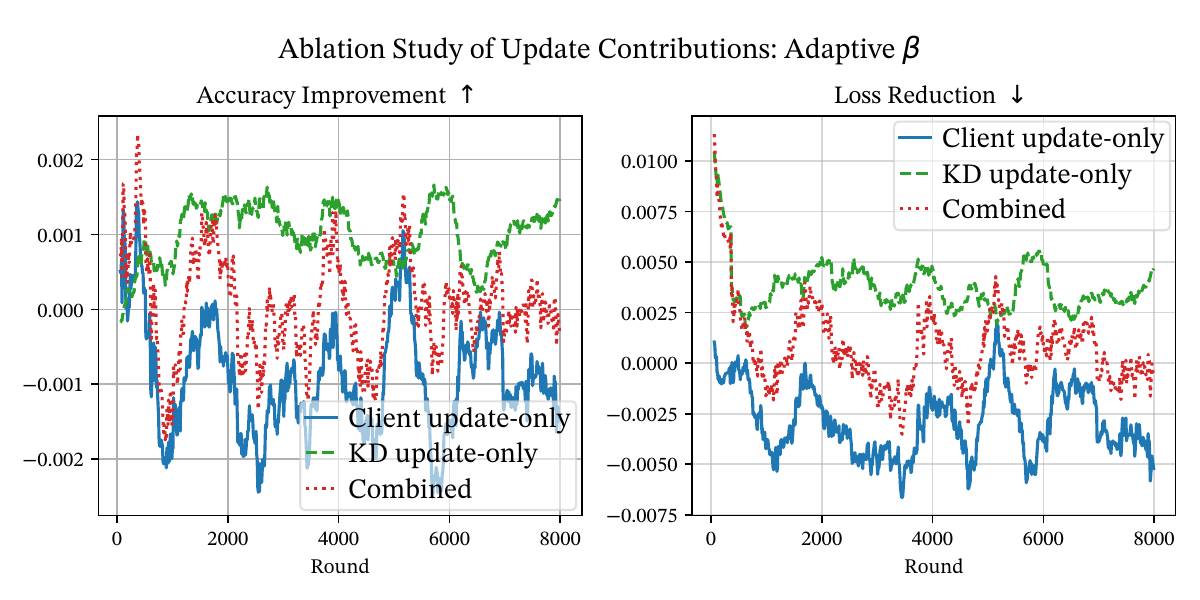}
  \vspace{-2em}
  \caption{Adaptive staleness-dependent schedule $\mybeta(\tau)$.}
  \label{fig:update_ablation_adaptive_beta}
\end{subfigure}
\hfill
\begin{subfigure}{0.48\textwidth}
  \centering
  \includegraphics[width=\textwidth]{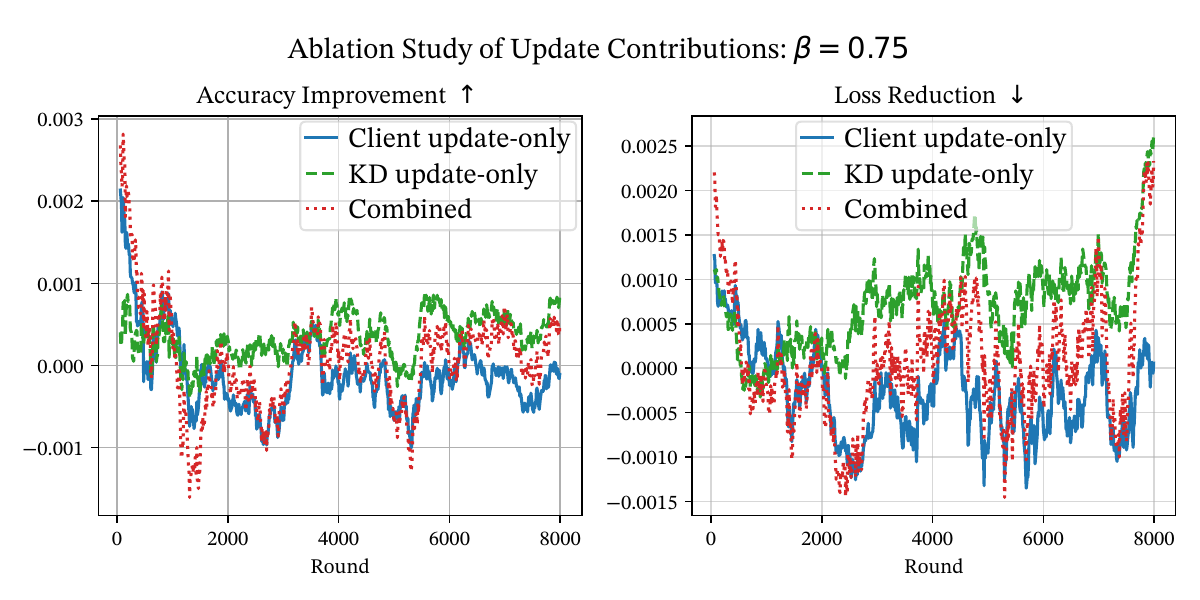}
  \vspace{-2em}
  \caption{Fixed mixing weight $\mybeta(\tau)=0.75$.}
  \label{fig:update_ablation_beta_075}
\end{subfigure}

\vspace{0.6em}

\begin{subfigure}{0.48\textwidth}
  \centering
  \includegraphics[width=\textwidth]{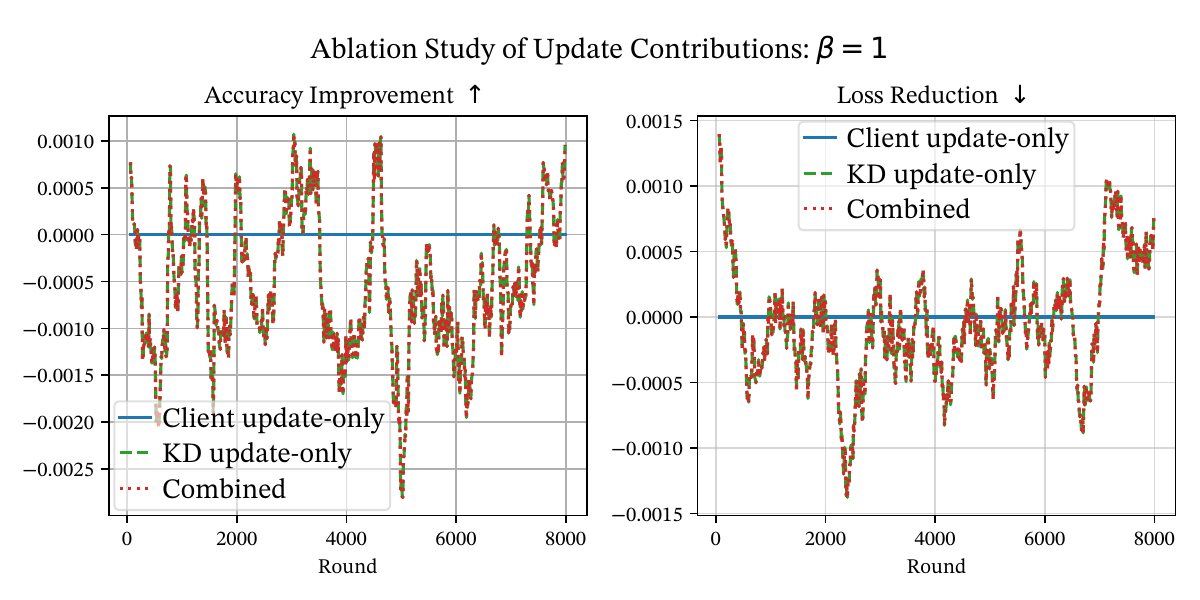}
  \vspace{-2em}
  \caption{Fixed mixing weight $\mybeta(\tau)=1$.}
  \label{fig:update_ablation_beta_1}
\end{subfigure}

\caption{\cifart{} update-component ablation under adaptive and fixed mixing-weight schedules.}
\label{fig:update_ablation_all}
\end{figure}

For completeness, we also provide the detailed breakdown for each mixing regime separately in \Cref{fig:update_ablation_all}.
Under the adaptive schedule, the combined update gives the strongest overall behavior, showing that the stale client update and the KD-revived update play complementary roles.
When $\mybeta$ is fixed to $0.75$ or $1$, the training dynamics deteriorate because the original client update is persistently underweighted or entirely ignored.
These results support the use of a staleness-aware schedule that balances parameter-space aggregation and KD-based revival rather than relying on a constant mixing rule.

\nips{
\subsection{Comparison of Different $\mybeta$ Functions}
\label{appx:beta_function_ablation}

We additionally compare several choices of the mixing function $\mybeta(\tau)$ that controls the interpolation between the stale client update and the KD-revived update in Eq.~(\ref{eq:ours_aggr}).
The purpose of this comparison is to evaluate how sensitive \ours{} is to the specific functional form of the staleness-dependent weighting.

In particular, we consider both adaptive schedules and constant baselines.
The adaptive schedules include several monotone families that increase the reliance on the KD component as staleness grows, while the constant baselines apply the same mixing weight at every staleness level.
Our main experiments use the \mbox{$1-\mathrm{cosine}$} schedule.
We find that the method is generally robust to the choice of $\mybeta$, and that even simple constant values around $0.5$ perform reasonably well.
However, adaptive schedules consistently provide the best overall trade-off, since they preserve more of the original client update when staleness is small and rely more on KD only when the update becomes sufficiently outdated.

\begin{figure}[t]
\centering
\begin{subfigure}[t]{0.48\textwidth}
    \centering
    \includegraphics[width=\linewidth]{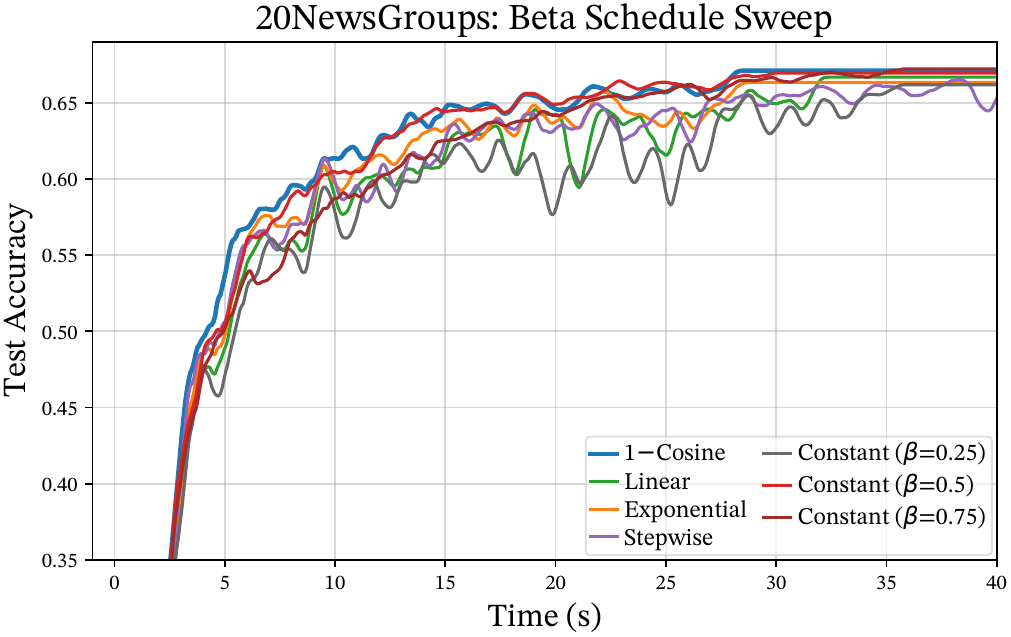}
    \caption{\news{}.}
\end{subfigure}\hfill
\begin{subfigure}[t]{0.48\textwidth}
    \centering
    \includegraphics[width=\linewidth]{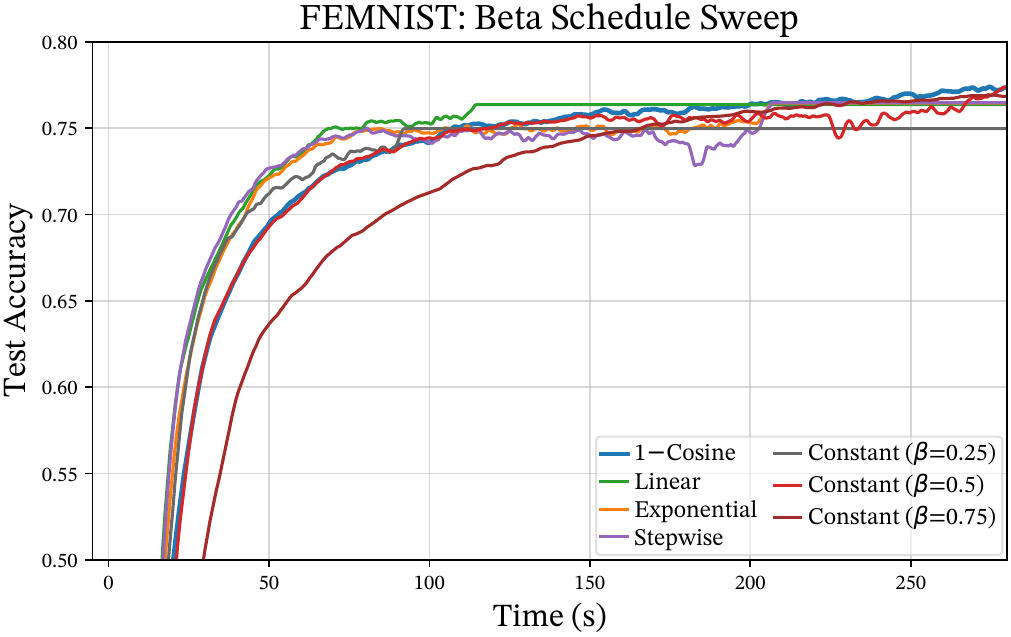}
    \caption{\femnist{}.}
\end{subfigure}
\caption{
\nips{
\textbf{Comparison of different $\mybeta(\tau)$ functions.}
Training curves under different adaptive and constant mixing schedules on \news{} and \femnist{}. The method is generally robust to the choice of $\mybeta$, while adaptive schedules provide the best overall trade-off by relying more on the raw client update at low staleness and shifting more weight to the KD-revived update as staleness increases.
}
}
\label{fig:beta_function_ablation}
\end{figure}

\Cref{fig:beta_function_ablation} reports the detailed comparison.
Overall, these results show that \ours{} is not highly sensitive to the exact choice of $\mybeta$, but benefits from using a schedule that increases smoothly with staleness rather than relying on a large constant value throughout training.
}

\nips{
\section{Server Memory Footprint of \ours{}}
\label{appx:memory_overhead}

\paragraph{Bounded additional server memory.}
\ours{} does not increase communication cost relative to standard asynchronous FL, but it does introduce bounded additional server-side memory. Beyond the current global model $\x[][\supp{t}]$, the server stores 
(i) a teacher buffer $\kdbuff$ containing $\kdbs$ recently received client models, and
(ii) the generator parameters $\genp$. Importantly, this overhead is controlled by fixed hyperparameters and does not grow with the total number of clients or the number of server rounds.

}

\nips{
\section{Privacy Discussion and Synthesized Samples}
\label{appx:privacy_discussion}

\ours{} does not introduce any additional client-side disclosures beyond the standard uploaded model updates already used in asynchronous FL. In particular, the server does not access raw client data, private labels, or any extra metadata beyond the regular model uploads also required by FedAvg-style methods. The pseudo-samples used by $\kdrevive$ are produced entirely on the server and serve only as an internal transfer set for knowledge distillation.

At the same time, we do not claim a new formal privacy guarantee. Our point is narrower and empirical: in our setting, the uploaded client updates are the result of batched local optimization over multiple steps and are further separated from the current server state by asynchronous delay. This communication interface appears sufficient to expose intermediate predictive structure that supports knowledge transfer, but not enough to synthesize realistic reconstructions of client images. Representative synthesized samples are shown in Fig.~\ref{fig:synth_samples_appendix}; they are typically low-fidelity and not visually interpretable as realistic client examples.

This observation is consistent with the role of the synthetic data in \ours{}: the samples are not intended to approximate the real client distribution or serve as a usable training dataset on their own, but rather to provide intermediate variables on which stale client models and the current server model can exchange predictive information through DFKD.

\begin{figure}[h]
    \centering
    \includegraphics[width=0.6\textwidth]{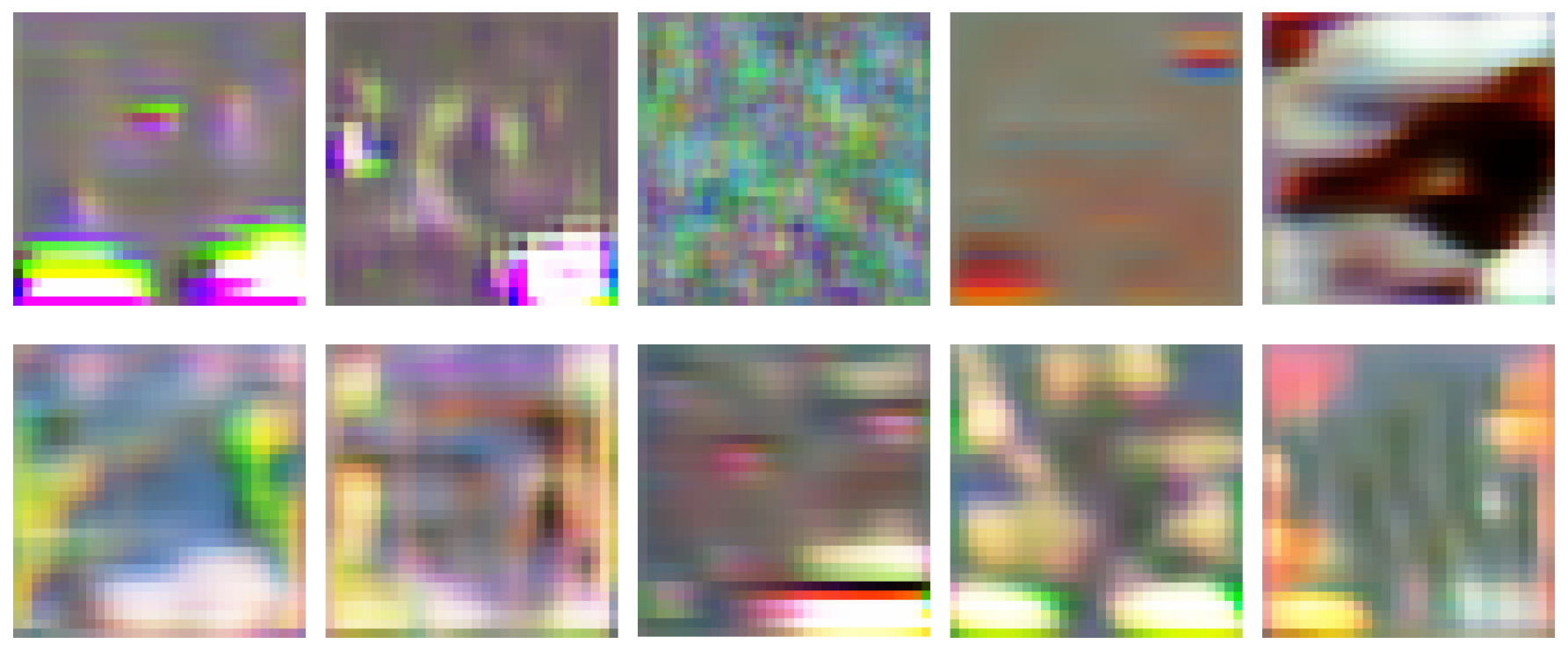}
    \caption{\nips{\textbf{Representative synthesized pseudo-samples used by \ours{}.} The samples are generated on the server for data-free knowledge distillation and are not intended to be realistic reconstructions of private client data. In our setting, regular asynchronous FL updates appear to provide enough signal for knowledge transfer, but not enough to recover visually realistic client images.}}
    \label{fig:synth_samples_appendix}
\end{figure}
}

\section{Experimental Details and Hyperparameters}
\label{appx:hyperparams}

\paragraph{Compute resources and runtime.}
All experiments are conducted on one of A100, V100, and H100 GPUs. Each experiment is run on a single GPU, and total runtime ranges from 2 to 20 hours. The dominant cost arises not from model training itself but from the sequential simulation of asynchronous federated learning with 1000 clients, which is required to realistically model client delays and staleness. This simulation introduces additional overhead compared to standard centralized or synchronous training, but does not require distributed multi-GPU execution. Memory requirements fit within standard GPU capacities (e.g., 16–80GB), depending on the model.

\paragraph{Data partitioning and seeds.}
We simulate $1000$ clients.
For all tasks except for $\femnist$, each client holds $350$ training examples, and client datasets are sampled using a non-iid Dirichlet partition. For $\femnist$ the clients follow natural user partition of the dataset.
We use $15\%$ of the full dataset for validation, $15\%$ for test.
All results are averaged over three random seeds. \camera{Our experiments assume benign clients. Extending \ours{} to address data or model poisoning is left for future work.}

\paragraph{Client optimization.}
Clients train with Adam ($\beta_1=0.9$), local batch size $32$, and $25$ local iterations per client update.
In asynchronous runs, $100$ clients are active concurrently out of $1000$ total clients. In synchronous runs, $100$ clients join a single round.

\noindent\textbf{Learning rates.}
We tune the client learning rate $\lrc$ and server learning rate $\lrs$ per method and dataset.
Table~\ref{tab:lr_table_updated_plain} reports the final choices. For \asyncdw{}, we use the same learning rates as \ours{}.
The ``Buffer / Window'' column reports the buffer size used by buffered async methods (\fedbuff{}, \fadas{}, \fedac{}) and the sliding-window size used by \fedfa{}; methods that aggregate every upload (\sync{}, \asyncdw{}, \oursdd{}, \ours{}) leave it empty.
Method-specific hyperparameters that are not learning rates (e.g., \fadas{} delay threshold $\tau_c$, \fedac{} momentum coefficients) are listed next.

\begin{table}[h]
\centering
\caption{Client and server learning rates used in each experiment, together with the buffer / window size for buffered async methods.}
\label{tab:lr_table_updated_plain}
\small
\begin{tabular}{lcccc}
\toprule
\textbf{Dataset} & \textbf{Method} & $\lrc$ & $\lrs$ & \textbf{Buffer / Window} \\
\midrule
\cifart{}  & \sync    & 0.0003 & 1.4    & -- \\
          & \fedbuff & 0.0003 & 0.05   & 2 \\
          & \fadas   & 0.0003 & 0.004  & 2 \\
          & \fedac   & 0.0003 & 0.0001 & 2 \\
          & \fedfa   & 0.0003 & 0.03   & 2 \\
          & \asyncdw & 0.0003 & 0.10   & -- \\
          & \oursdd  & 0.0003 & 0.20   & -- \\
          & \ours    & 0.0003 & 0.10   & -- \\
\midrule
\cifarh{}  & \sync    & 0.0003 & 1.4    & -- \\
          & \fedbuff & 0.0003 & 0.10   & 3 \\
          & \fadas   & 0.0003 & 0.003  & 3 \\
          & \fedac   & 0.0003 & 3e-5   & 3 \\
          & \fedfa   & 0.0003 & 0.03   & 2 \\
          & \asyncdw & 0.0003 & 0.10   & -- \\
          & \oursdd  & 0.0001 & 0.30   & -- \\
          & \ours    & 0.0003 & 0.10   & -- \\
\midrule
\femnist{} & \sync    & 0.001  & 1.4    & -- \\
          & \fedbuff & 0.001  & 0.025  & 1 \\
          & \fadas   & 0.001  & 0.003  & 1 \\
          & \fedac   & 0.001  & 0.0001 & 2 \\
          & \fedfa   & 0.001  & 0.015  & 2 \\
          & \asyncdw & 0.001  & 0.05   & -- \\
          & \oursdd  & 0.001  & 0.10   & -- \\
          & \ours    & 0.001  & 0.05   & -- \\
\midrule
\news{}    & \sync    & 0.003 & 1.0    & -- \\
          & \fedbuff & 0.001 & 0.1    & 3 \\
          & \fadas   & 0.001 & 0.005  & 3 \\
          & \fedac   & 0.001 & 0.0003 & 3 \\
          & \fedfa   & 0.001 & 0.03   & 2 \\
          & \asyncdw & 0.001 & 0.1    & -- \\
          & \oursdd  & 0.001 & 0.1    & -- \\
          & \ours    & 0.001 & 0.1    & -- \\
\bottomrule
\end{tabular}
\end{table}

\paragraph{\fadas{} hyperparameters.}
\fadas{} uses an AMSGrad-style server optimizer with $\beta_1=0.9$, $\beta_2=0.99$, and $\epsilon=10^{-8}$.
The delay threshold $\tau_c$ that gates the adaptive step is set to $\tau_c=3$ on \cifart{}, \cifarh{}, and \femnist{}, and $\tau_c=2$ on \news{}.

\paragraph{\fedac{} hyperparameters.}
\fedac{} uses a Nadam-style server optimizer with prospective momentum coefficients $\beta_1=0.6$, $\beta_2=0.9$, and $\epsilon=10^{-8}$.
Following the original formulation, client weights are derived from the cosine similarity between the global model drift and each client's pseudo-gradient (clipped at zero and normalized).
We adopt the Basic-WTG-APWM variant and drop the client-side gradient correction so that all methods share the same client training procedure for a fair comparison.

\paragraph{\fedfa{} hyperparameters.}
\fedfa{} aggregates with a fixed-size sliding window of $K=2$ uploads on every dataset; the global model is refreshed on every upload once the window is full, rather than waiting for $K$ \emph{new} updates as in \fedbuff{}.

\vspace{0.5em}
\noindent\textbf{KD augmentation settings.}
Both KD-based variants use a teacher buffer of size $\kdbs=8$ and a KD batch size of $32$.

\paragraph{\oursdd{} hyperparameters.}
We run $10$ distillation steps per server update, and fix the mixing coefficient to $\beta=0.5$.
Table~\ref{tab:simplekd_table_plain} reports the student learning rate used inside KD and the distillation temperature.

\begin{table}[h]
\centering
\caption{Hyperparameters for \oursdd{}. Shared: $10$ distillation steps per server update, distillation batch size $32$, teacher buffer size $\kdbs=8$, and $\beta=0.5$ (constant for \oursdd{}).}
\label{tab:simplekd_table_plain}
\small
\begin{tabular}{lcc}
\toprule
\textbf{Dataset} & \textbf{KD learning rate} & \textbf{Distillation temperature} \\
\midrule
\cifart{}  & 0.0003 & 2.0 \\
\cifarh{}  & 0.0001 & 2.0 \\
\femnist{} & 0.0003 & 2.0 \\
\news{}    & 0.0001 & 2.0 \\
\bottomrule
\end{tabular}
\end{table}

\paragraph{\ours{} hyperparameters.}
We use synthesis batch size $64$ and KD batch size $32$.
Per server update, we perform $10$ KD steps and update the generator for $2$ steps before the meta-learning step, once every $10$ server updates.
The KD temperature is fixed to $\text{T}=1$.
For the synthesis loss in Eq.~(\ref{eq:lsynth_combined_twolines}), we fix the adversarial and target alignment weights to $\alpha_{\text{adv}}=0.1$ and $\alpha_{\text{target}}=1.0$, and set the feature matching weight $\alpha_{\text{feature}}$ per dataset (Table~\ref{tab:dfkd_table_plain}).
For the aggregation mixing, \ours{} uses the adaptive schedule described in the main text.

\begin{table}[h]
\centering
\caption{Dataset-specific DFKD hyperparameters for \ours{}. Shared: synthesis batch size $64$, KD batch size $32$, teacher buffer size $8$, KD temperature $\text{T}=1$, $2$ generator steps every $10$ server updates, and $10$ KD steps per server update. Learning rates (LR) are presented in the table.}
\label{tab:dfkd_table_plain}
\small
\begin{tabular}{lcccc}
\toprule
\textbf{Dataset} &
\textbf{KD LR} &
\textbf{Generator LR} &
\textbf{Latent noise LR} &
$\alpha_{\text{feature}}$ \\
\midrule
\cifart{}  & 1e-4 & 0.003 & 0.001   & 0.003 \\
\cifarh{}  & 3e-5 & 0.003 & 0.001   & 0.0003 \\
\femnist{} & 1e-4 & 0.01  & 0.003  & 0.3 \\
\news{}    & 3e-4 & 0.01  & 0.1     & 0.0001 \\
\bottomrule
\end{tabular}
\end{table}

\paragraph{Feature statistics used in $\Lfeature$.}
The feature loss in Eq.~(\ref{eq:lsynth_combined_twolines}) matches architecture-specific intermediate statistics via $\phi(\cdot)$.
For the vision backbones (ResNet-18 on \cifart{}/\cifarh{} and the CNN on \femnist{}), we follow standard DFKD practice and use BatchNorm statistics.
For the transformer model (\news{} with T5-small), we use hidden-states at the position of classification head (first token).

\subsection{$\beta$ Scheduling Families}
\label{appx:beta_families}
\begin{figure}[h]
\centering
\includegraphics[width=0.5\textwidth]{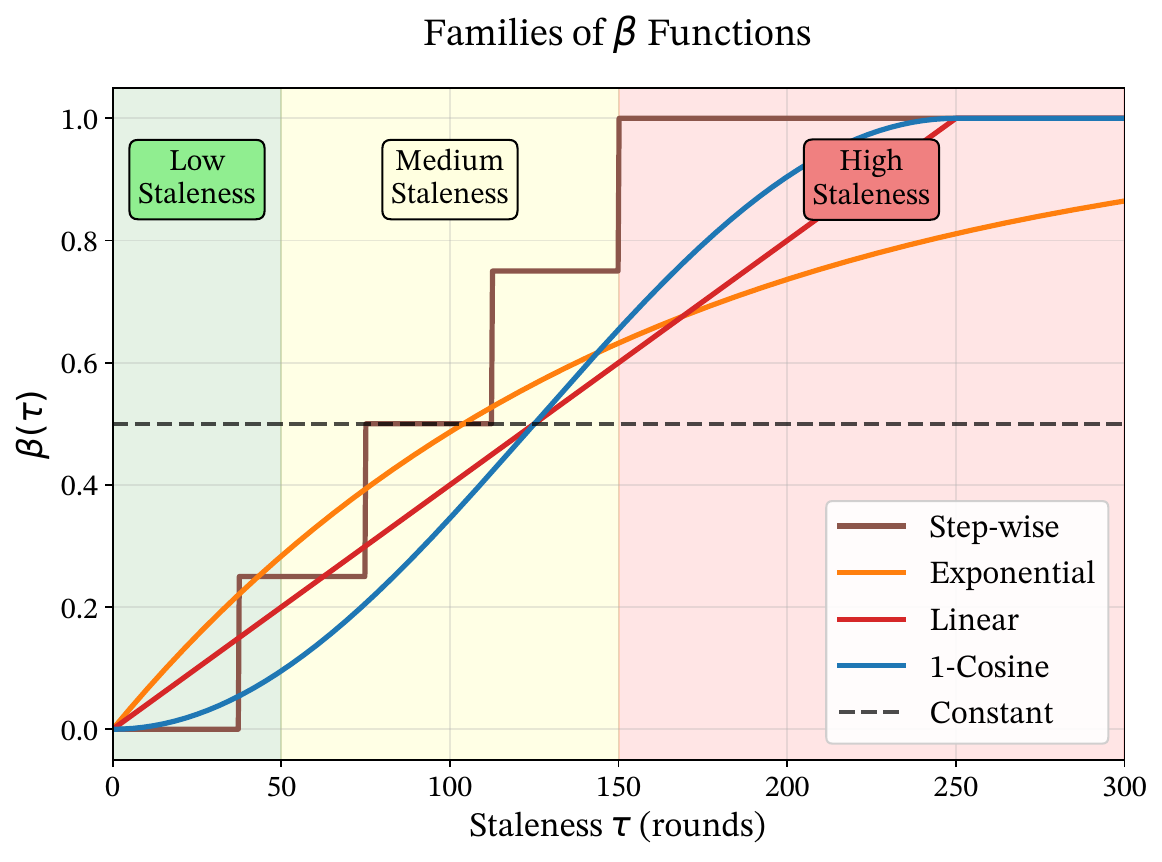}
\caption{Different candidate $\beta$ scheduling functions controlling the mix between parameter aggregation and DFKD updates. \mbox{$1-\mathrm{cosine}$} schedule is chosen, as it empirically shows the best performance on \femnist{} and \news{}.}
\label{fig:beta_families}
\end{figure}

We consider several candidate monotone scheduling families for $\mybeta$, which determines how quickly \ours{} shifts from parameter-space aggregation to KD-dominated aggregation as staleness increases. \Cref{fig:beta_families} visualizes the families we considered. Unless stated otherwise, we use the \mbox{$1-\mathrm{cosine}$} schedule in all experiments. The choice is explained in detail in \apdx~\ref{appx:beta_function_ablation}.

\end{document}